\documentclass[10pt,twocolumn,letterpaper]{article}

\usepackage[pagenumbers]{cvpr} %

\usepackage{graphicx}
\usepackage{amsmath}
\usepackage{amssymb}
\usepackage{booktabs}
\usepackage{enumitem}

\usepackage[pagebackref,breaklinks,colorlinks]{hyperref}

\usepackage[capitalize]{cleveref}
\crefname{section}{Sec.}{Secs.}
\Crefname{section}{Section}{Sections}
\Crefname{table}{Table}{Tables}
\crefname{table}{Tab.}{Tabs.}

\renewcommand{\paragraph}[1]{\noindent {\bf #1}}

\begin{document}

\title{Shape, Pose, and Appearance from a Single Image\\via Bootstrapped Radiance Field Inversion}

\author{Dario Pavllo\textsuperscript{1,2}$^*$ \qquad David Joseph Tan\textsuperscript{2} \qquad Marie-Julie Rakotosaona\textsuperscript{2} \qquad Federico Tombari\textsuperscript{2,3}\vspace{2mm}\\
\textsuperscript{1}ETH Zurich \qquad \textsuperscript{2}Google \qquad \textsuperscript{3}TU Munich\vspace{3mm}
}

\twocolumn[{%
\renewcommand\twocolumn[1][]{#1}%
\maketitle
\begin{center}
    \centering
    \vspace{-10.5mm}
    \includegraphics[width=\textwidth, trim={0pt 17pt 0pt 2pt}, clip]{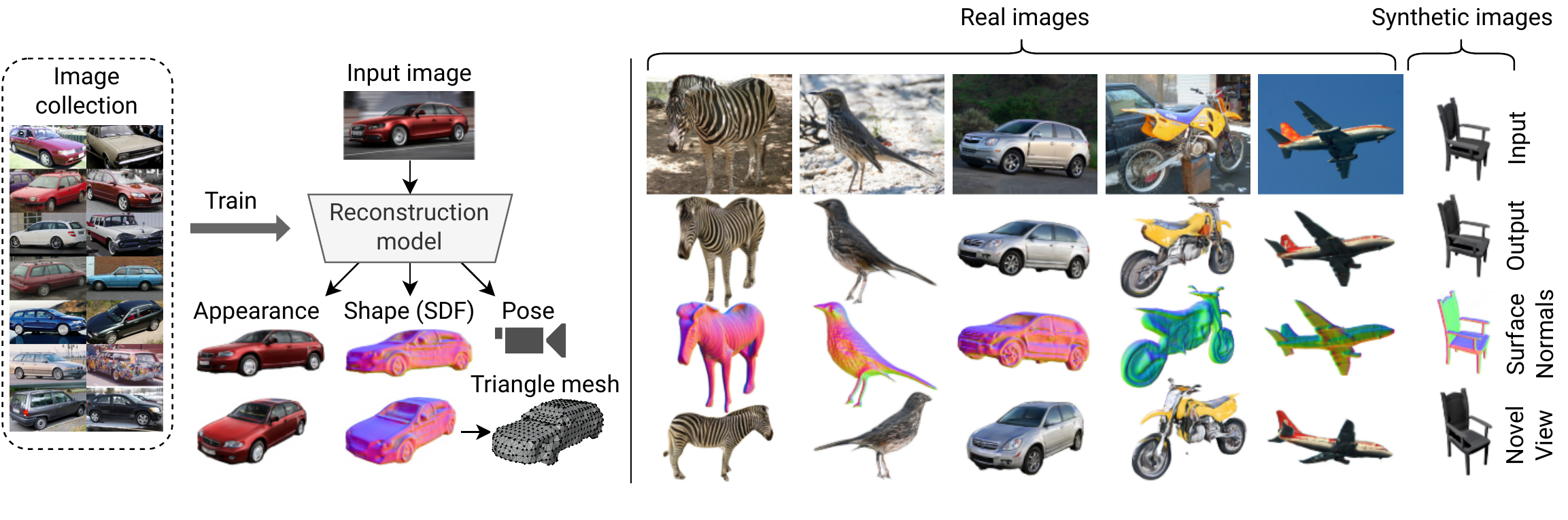}
    \captionof{figure}{Given a collection of 2D images representing a specific category (\eg cars), we learn a model that can fully recover shape, pose, and appearance from a single image, without leveraging multiple views during training. The 3D shape is parameterized as a signed distance function (SDF), which facilitates its transformation to a triangle mesh for further downstream applications. %
    }
    \label{fig:teaser}
\end{center}%
}]

\def\thefootnote{$*$}\footnotetext{Work done during an internship at Google.}\def\thefootnote{\arabic{footnote}}

\begin{abstract}
   Neural Radiance Fields (NeRF) coupled with GANs represent a promising direction in the area of 3D reconstruction from a single view, owing to their ability to efficiently model arbitrary topologies. Recent work in this area, however, has mostly focused on synthetic datasets where exact ground-truth poses are known, and has overlooked pose estimation, which is important for certain downstream applications such as augmented reality (AR) and robotics. We introduce a principled end-to-end reconstruction framework for natural images, where accurate ground-truth poses are not available. Our approach recovers an SDF-parameterized 3D shape, pose, and appearance from a single image of an object, without exploiting multiple views during training. More specifically, we leverage an unconditional 3D-aware generator, to which we apply a hybrid inversion scheme where a model produces a first guess of the solution which is then refined via optimization. Our framework can de-render an image in as few as 10 steps, enabling its use in practical scenarios. We demonstrate state-of-the-art results on a variety of real and synthetic benchmarks.
\end{abstract}

\section{Introduction}
\label{sec:introduction}

We focus on single-view 3D reconstruction, where the goal is to reconstruct shape, appearance, and camera pose from a single image of an object (\autoref{fig:teaser}). Such a task has applications in content creation, augmented \& virtual reality (AR/VR), robotics, and is also interesting from a scientific perspective, as most neural architectures cannot reason about 3D scenes. %
As humans, we learn \emph{object priors}, abstract representations that allow us to imagine what a partially-observed object would look like from other viewpoints. Incorporating such knowledge into a model would enable higher forms of 3D reasoning. While early work on 3D reconstruction has focused on exploiting annotated data \cite{girdhar2016learning, zhu2017rethinking, wu2017marrnet, yang20173d, hane2017hierarchical}, \eg ground-truth 3D shapes or multiple 2D views, more recent work has relaxed the assumptions required by the task. In particular, there has been effort in learning this task from single-view collections of images depicting a specific category \cite{kanazawa2018cmr, goel2020ucmr, li2020umr} (\eg a dataset of cars), and we also follow this line of work.

Most established 3D representations in the single-view reconstruction literature are based on deformable triangle meshes \cite{kanazawa2018cmr, goel2020ucmr, li2020umr}, although Neural Radiance Fields (NeRF) \cite{mildenhall2020nerf, barron2021mipnerf} have recently become more prominent in the broader 3D vision community owing to their ability to efficiently model arbitrary topologies. These have been combined with GANs \cite{goodfellow2014gan} for unconditional 3D generation tasks \cite{chan2021pigan, chan2022eg3d, niemeyer2021giraffe, xue2022giraffehd}, as they produce more perceptually pleasing results. %
There has also been work on combining the two in the single-view reconstruction task, \eg Pix2NeRF \cite{cai2022pix2nerf}, which is however demonstrated on simple settings of faces or synthetic datasets where perfect ground-truth poses are available. Furthermore, there has been less focus overall on producing an end-to-end reconstruction system that additionally tackles \emph{pose estimation} (beyond simple settings), which is particularly important for AR applications. In our work, we bridge this gap by proposing a more general NeRF-based end-to-end reconstruction pipeline that tackles both reconstruction and pose estimation, and demonstrate its broader applicability to natural images where poses cannot be accurately estimated. We further characterize the problem by comparing encoder-based approaches (the majority of methods in the single-view reconstruction literature) to inversion-based approaches (which invert a generator via optimization), and show that the latter are more suited to real datasets without accurate ground-truth poses.

Motivated by this, we propose a \emph{hybrid GAN inversion} technique for NeRFs that can be regarded as a compromise between the two: an encoder produces a first guess of the solution (\emph{bootstrapping}), which is then refined via optimization. We further propose a series of technical contributions, including: \emph{(i)} the adoption of an SDF representation \cite{yariv2021volsdf} to improve the reconstructed surfaces and facilitate their conversion to triangle meshes, \emph{(ii)} regularizers to accelerate inversion, and \emph{(iii)} the addition of certain equivariances in the model architecture to improve generalization. We show that we can invert an image in as few as 10 optimization steps, making our approach usable even in constrained scenarios. Furthermore, we incorporate a principled pose estimation framework \cite{wang2019nocs} that frames the problem as a regression of a canonical representation followed by Perspective-\emph{n}-Point (PnP), and show that it boosts pose estimation accuracy without additional data assumptions.
We summarize our main contributions as follows:
\vspace{-0.5mm}
\begin{itemize}[leftmargin=*, itemsep=-2pt]
    \item We introduce an end-to-end single-view 3D reconstruction pipeline based on NeRFs. In this setting, we successfully demonstrate 360$^\circ$ object reconstruction from natural images under the CMR \cite{goel2020ucmr} benchmark.
    \item We propose a hybrid inversion scheme for NeRFs to accelerate the reversal of pre-trained 3D-aware generators.
    \item Inspired by the literature on pose estimation, we propose a principled PnP-based pose estimator that leverages our framework and does not require extra data assumptions.
\end{itemize}
\vspace{-0.5mm}

To validate our contributions, we obtain state-of-the-art results on both real/synthetic benchmarks. Furthermore, to our knowledge, we are the first to demonstrate NeRF-based reconstruction on \emph{in-the-wild} datasets such as ImageNet.

We release our code and pretrained models at {\small\url{https://github.com/google-research/nerf-from-image}}.
\section{Related work}
\label{sec:related-work}

\paragraph{Inverse rendering and scene representations.} Although 3D reconstruction is an established task, the representations and supervision methods used to tackle this problem have evolved throughout the literature. Early approaches have focused on reconstructing shapes using 3D supervision, adopting voxel grids \cite{girdhar2016learning, zhu2017rethinking, wu2017marrnet, yang20173d, hane2017hierarchical}, point clouds \cite{fan2017point}, or SDFs \cite{park2019deepsdf}, and require synthetic datasets where ground-truth 3D shapes are available. The introduction of differentiable rendering \cite{loper2014opendr, kato2018n3mr, liu2019softras, chen2019dibr, chen2021dibrpp} has enabled a new line of work that attempts to reconstruct shape and texture from single-view datasets, leveraging triangle mesh representations \cite{kanazawa2018cmr, goel2020ucmr, li2020umr, chen2019dibr, bhattad2021viewgen, zhang2022meshinversion, bmvcHendersonF18}.
Each 3D representation, however, comes with its own set of trade-offs. For instance, voxels do not scale efficiently with resolution, while triangle meshes are efficient but struggle with arbitrary topologies (most works deform a sphere template). In recent developments, implicit representations encode a 3D scene as the weights of an MLP that can be queried at specific coordinates, which allows them to model arbitrary topologies using lightweight networks. In such a setting, there has been work on 3D reconstruction using implicit SDFs \cite{lin2020sdfsrn, duggal2022tars} as well as neural radiance fields (NeRF) \cite{mildenhall2020nerf, barron2021mipnerf}. %
Finally, some works incorporate additional structural information into 3D representations, \eg \cite{yao2022lassie} reconstructs articulated shapes using skeleton priors, \cite{chen2021dibrpp, wimbauer2022derendering} disentangle albedo from reflectance, and \cite{xu2022sinnerf} uses depth cues. These techniques are orthogonal to ours and may positively benefit each other.

\paragraph{NeRF-based reconstruction.} The standard use-case of a NeRF is to encode a single scene given multiple 2D views and associated camera poses, which does not necessarily lead to learned shared representations. There have however been attempts at learning an object prior by training such models on a category-specific dataset (\eg a collection of cars). For instance, \cite{rebain2022lolnerf, jang2021codenerf} train a shared NeRF backbone conditioned on a learned latent code for each object instance. \cite{yu2020pixelnerf} tackles reconstruction conditioned on an image encoder, although it requires multiple ground-truth views for supervision and does not adopt an adversarial setting, thereby relying on accurate poses from synthetic datasets and leading to blurry results. \cite{cai2022pix2nerf, mi2022im2nerf} adopt an adversarial setting and only require a single view during training, but they focus on settings with simple pose distributions. Finally, there has been work on using diffusion models \cite{ho2020denoising, watson2022nvsdiffusion} and distillation \cite{ramirez2021unsupervised} for novel-view synthesis, though such methods do not explicitly recover a 3D surface.

\paragraph{Encoder- \emph{vs} inversion-based methods.} Most aforementioned methods can be categorized as \emph{encoder-based}, where a 2D ConvNet encodes the input image into a latent representation, then decoded into a 3D scene. %
This paradigm is analogous to an autoencoder, and therefore requires some form of pixel-level loss between predicted and input images. While this is appropriate for synthetic datasets with exact poses, it leads to blurry or distorted results when such poses are inaccurate (\ie the case in natural images). %
Following the 2D GAN inversion literature \cite{xia2022ganinversion}, there has been work on applying inversion methods to 3D reconstruction, where the goal is to leverage a pretrained unconditional GAN and find the latent code that best fits the input image via optimization. Since unconditional GANs tend to be more robust to inaccurate poses (as they mostly rely on the overall pose distribution as opposed to the pose of each image), we argue that inversion-based approaches are better suited to natural images. As part of our work, we characterize this phenomenon experimentally. 3D GAN inversion has been applied to untextured shapes \cite{zhang2021unsupervised, duggal2022mending}, textured triangle meshes \cite{zhang2022meshinversion}, and its use with NeRF-based approaches is suggested in \cite{chan2021pigan, cai2022pix2nerf, chan2022eg3d}, although it is not their focus.

\paragraph{Our work.} We propose a \emph{hybrid} inversion paradigm, where an encoder produces a first guess of the latent representation and pose (\emph{bootstrapping}), and these are then refined for a few iterations via optimization. Although \cite{duggal2022mending} introduce a similar idea, they focus on shape completion from LiDAR data, whereas we focus on shape, pose, and appearance prediction from an image. Under our setting, Pix2NeRF \cite{cai2022pix2nerf} provides a proof-of-concept of refinement using such a method, but it is still trained using an encoder-based paradigm and is thus affected by the aforementioned issues. By contrast, we propose a principled end-to-end hybrid reconstruction approach that takes full advantage of an unconditional generator and can also optimize with respect to pose (unlike \cite{chan2021pigan, cai2022pix2nerf, chan2022eg3d}), a task that requires a suitable pose parameterization. We also mention that \cite{zhang2021imagegan} propose a similar idea to bootstrapping (without inversion), but they adopt a 2D image generator as opposed to a 3D-aware one, which does not fully disentangle pose from appearance.  

\paragraph{Unconditional generation.} Since inversion-based approaches rely on a pretrained generator, we briefly discuss recent architectures for this task. \cite{pavllo2020convmesh, pavllo2021textured3dgan, henderson2020leveraging} learn to generate triangle meshes and textures using 2D supervision from single-view collections of natural images. \cite{chan2021pigan} learns this task using NeRFs, although it suffers from the high computational cost of MLP-based NeRFs. \cite{niemeyer2021giraffe, xue2022giraffehd, schwarz2020graf, or2022stylesdf, gu2021stylenerf, sitzmann2019srn, chan2022eg3d} incorporate both 2D and 3D components as a trade-off between 3D consistency and efficiency.
Finally, \cite{gao2022get3d} proposes an approach to train a NeRF-based generator whose outputs can be distilled into triangle meshes. 
The generator used in our work leverages an EG3D-like backbone \cite{chan2022eg3d}. %

\section{Method}
\label{sec:method}
We now present our single-view reconstruction approach. We break down our method into three main steps. \emph{(i)} Initially, we train an unconditional generator following the literature on 3D-aware GANs \cite{chan2021pigan, chan2022eg3d}, where a NeRF-based generator $\mathbf{G}$ is combined with a 2D image discriminator. This framework requires minimal assumptions, namely 2D images and the corresponding pose distribution. %
We further apply a series of technical improvements to the overall framework in order to positively impact the subsequent reconstruction step, as explained in \autoref{sec:method-unconditional-generation}. \emph{(ii)} We freeze $\mathbf{G}$ and train an image encoder $\mathbf{E}$ that jointly estimates the pose of the object as well as an initial guess of its latent code (\emph{bootstrapping}). For pose estimation, we adopt a principled approach that predicts a canonical map \cite{wang2019nocs} in screen space followed by a Perspective-\emph{n}-Point (PnP) algorithm. We explain these steps in \autoref{sec:method-latent-regressor}. Finally, \emph{(iii)} we refine the pose and latent code for a few steps via gradient-based optimization (\emph{hybrid inversion}), as described in \autoref{sec:method-hybrid-inversion}.

\paragraph{Requirements.} For training, our method requires a category-specific collection of images, along with segmentation masks for datasets with a background (we use an off-the-shelf segmentation model, PointRend \cite{kirillov2020pointrend}), which we use to pre-segment the images. An approximate pose distribution must also be known. For inference, only a single, unposed input image is required.

\subsection{Unconditional generator pre-training}
\label{sec:method-unconditional-generation}

\begin{figure}[h]
\centering
  \includegraphics[width=\linewidth, trim={25pt 3pt 0pt 0pt}, clip]{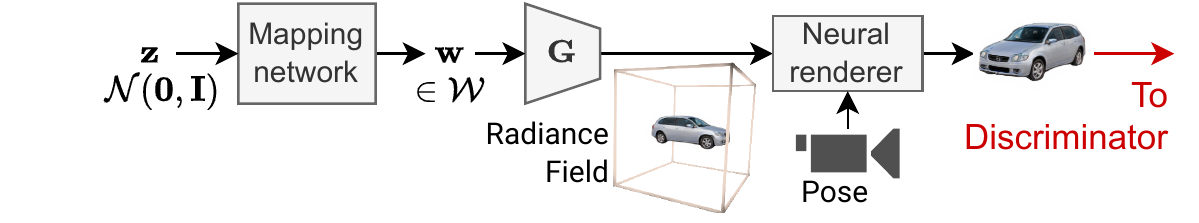}
   \caption{Unconditional generation framework.}
\label{fig:unconditional-generation}
\end{figure}

We adopt EG3D \cite{chan2022eg3d} as a starting point for the backbone of our generator. It consists of a \emph{mapping network} that maps a prior $\mathbf{z} \sim \mathcal{N}(\mathbf{0}, \mathbf{I})$ to a latent code $\mathbf{w} \in \mathcal{W}$, the latter of which is plugged into a StyleGAN2 generator \cite{karras2020styleganv2}. The output feature map is then split into three orthogonal planes (\emph{xy}, \emph{xz}, \emph{yz}), which are queried at specific coordinates via bilinear sampling. The resulting features are finally summed and plugged into a tiny MLP (\emph{triplanar decoder}) to output the values of the radiance field (density and color). The generator $\mathbf{G}$ is trained using a GAN framework where the discriminator takes 2D renderings as input. %
We apply some adjustments to the triplanar decoder and training objective, including the ability to model view-dependent effects as well as improvements to the adaptive discriminator augmentation (ADA) technique \cite{karras2020ada}, which is used on small datasets (see Appendix \ref{sec:appendix-implementation-details}). In the next paragraphs, we focus on the changes that are central to our reconstruction approach.

\paragraph{SDF representation.} We found it beneficial to parameterize the object surface as a signed distance function (SDF), as opposed to the standard volume density parameterization adopted in EG3D \cite{chan2022eg3d}. In addition to an empirical advantage (\autoref{sec:results}), SDFs facilitate the extraction of the surface and its subsequent conversion to other representations (\eg triangle meshes), since they provide analytical information about surface boundaries and normals. SDFs have already been explored in unconditional generators \cite{or2022stylesdf} and in the broader NeRF literature \cite{oechsle2021unisurf, yariv2021volsdf, wang2021neus, yu2022monosdf}, but less so in the single-view reconstruction setting.
We follow VolSDF \cite{yariv2021volsdf}, in which the volume density $\sigma(\mathbf{x})$ is described as:
\setlength{\abovedisplayskip}{3pt}
\setlength{\belowdisplayskip}{3pt}
\begin{equation}
    \label{eq:sdf}
    \sigma{(\mathbf{x}}) = (1/\alpha)\,\Psi_\beta(-d(\mathbf{x}))\, ,
\end{equation}
where $\mathbf{x}$ are the query coordinates, $d(\mathbf{x})$ is the SDF (\ie the output of the generator), and $\Psi_\beta$ is the cumulative density function (CDF) of the Laplace distribution with scale $\beta$ and zero mean. $\alpha, \beta > 0$ are learnable parameters. We also incorporate an Eikonal loss to encourage the network to approximate a valid distance function:
\setlength{\abovedisplayskip}{3pt}
\setlength{\belowdisplayskip}{3pt}
\begin{equation}
    \mathcal{L}_\text{Eikonal} = \mathbb{E}_\mathbf{x}[(\lVert \nabla_\mathbf{x} d(\mathbf{x}) \rVert - 1)^2].
\end{equation}
We efficiently approximate the expectation using stratified sampling across the bounding volume of the scene, and employ a custom bilinear sampling implementation in the triplanar encoder which supports double differentiation w.r.t.\ the input query points. Furthermore, we initialize the SDF to a unit sphere via pre-training. Implementation details can be found in the Appendix \ref{sec:appendix-implementation-details}.

\paragraph{Removing super-resolution network.} In \cite{chan2022eg3d}, the rendered image is further processed through a super-resolution network, which increases its resolution and corrects for any distribution mismatch at the expense of 3D consistency. %
Since we aim to address fully 3D-consistent reconstruction instead of a more relaxed novel-view-synthesis task, we remove this component and feed the rendered image directly through the discriminator. This choice also makes it easier to fairly compare our approach to existing work.

\begin{figure}[t]
\centering
  \includegraphics[width=\linewidth, trim={28pt 0pt 10pt 5pt}, clip]{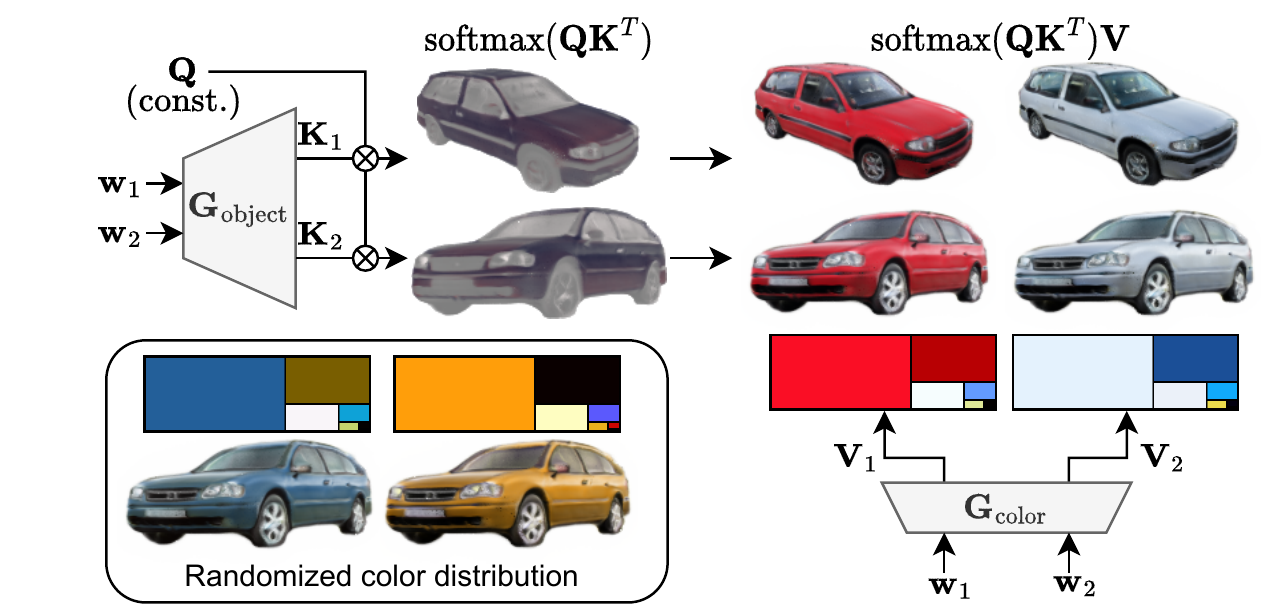}
   \caption{Illustration of our color mapping technique with two objects generated by two different latent codes $\mathbf{w}_1$ and $\mathbf{w}_2$. The object generator models a latent radiance field of keys $\mathbf{K}$ (each of which represents a semantic embedding at a specific spatial position), which are multiplied with a fixed set of queries $\mathbf{Q}$ (\ie learned prototype embeddings for each ``semantic channel'') and processed through a softmax to produce a probability distribution across these semantic channels, whose meaning is learned. In the case of cars, the learned semantic channels include body, headlights, wheels, and reflections. In the image, we show a rendering of the result of this operation in false colors, where the weight of one of the classes (car body) is highlighted. Finally, the latter is multiplied with the values $\mathbf{V}$ (color distribution, \ie a color for each semantic channel) produced by another module (color network), resulting in the final RGB colors. While during training the same latent code goes into both networks so as to learn the correct data distribution, at inference we can split it to swap the color distribution among different object identities (\textbf{top-right}) or randomize it entirely (\textbf{bottom-left}).}
\label{fig:color-mapping}
\end{figure}

\paragraph{Attention-based color mapping.} A robust 3D reconstruction technique should be as much as possible equivariant to certain transformations in order to improve generalization on unseen data. These include geometric transformations (\eg a 2D translation in the input image should be reflected in the 3D pose, which motivates our principled pose estimation technique in \autoref{sec:method-latent-regressor}) as well as color transformations, \eg changing the hue of an object (an image of a red car into that of a white car) should result in an equivalent change in the radiance field. As an extreme example, without such an equivariance incorporated in the architecture, a model trained on a dataset of red cars will not generalize to one of white cars. This motivates us to disentangle the color distribution from the identity (pseudo-``semantics'') of the generated objects, as shown in \autoref{fig:color-mapping}.

Our formulation is a soft analogy to UV mapping, where the lookup is done through an attention mechanism instead of texture coordinates.
This approach additionally provides simple manipulation capabilities (see \autoref{fig:color-mapping}). %
A useful property of our formulation is that the color mapping operator is linear w.r.t.\ the colors. It can be applied either \emph{before} (in the radiance field sampler) or \emph{after} the rendering operation (in the rendered multi-channel ``semantic image''), since the rendering operation is also linear w.r.t.\ the colors. In a reconstruction scenario, this allows the end user to efficiently reproduce the color distribution of the input image with a single rendering pass. In \autoref{sec:results} we show that, in addition to the useful manipulation properties, this module leads to an empirical advantage in the reconstruction task. %

\paragraph{Path Length Regularization revisited.} Initially proposed in StyleGAN2 \cite{karras2020styleganv2}, this regularizer encourages the mapping between the latent space $\mathcal{W}$ and the output space $\mathcal{Y}$ to be orthogonal, which facilitates inversion (recovering the latent code $\mathbf{w}$ corresponding to a certain image via optimization). This is achieved by applying a gradient penalty to the Jacobian $\partial g(\mathbf{w}) / \partial \mathbf{w}$. The use of path length regularization on the full backbone, however, is prohibitively expensive as this term requires double differentiation, and this feature was dropped in EG3D \cite{chan2022eg3d}. We propose to reinstate a more efficient variant of this regularizer which computes the path length penalty up to the three orthogonal planes, leaving the triplanar decoder unregularized. We find that this compromise provides the desired benefits without a significant added computational cost, as the main bottleneck is represented by the triplanar decoder, and enables us to greatly increase the learning rate during the inversion process (and in turn reduce the number of iterations).

\subsection{Bootstrapping and pose estimation}
\label{sec:method-latent-regressor}
Given a pretrained generator, it is in principle possible to invert it using one of the many techniques described in the literature for 2D images \cite{roich2022pti}, which usually involve minimizing some pixel-level loss (\eg L1 or VGG) w.r.t.\ the input latent code. For the 3D case, the minimization needs to be carried out over both the latent code and camera pose. In practice, however, recovering the camera pose is a highly non-convex problem that can easily get stuck in local minima. %
It is also crucial that the initial pose is ``good enough'', otherwise the latent code will converge to a degenerate solution. Therefore, most approaches \cite{chan2022eg3d, cai2022pix2nerf} initialize the pose using an off-the-shelf pose estimator and only carry out the optimization w.r.t.\ the latent code. Moreover, existing approaches start from an average or random latent code \cite{chan2022eg3d, zhang2022meshinversion}, resulting in a slow convergence (often requiring hundreds of steps), which makes these methods less applicable to real-time scenarios. This motivates our hybrid inversion scheme, where an encoder produces a first guess of the latent code and pose, and these are \emph{both} refined for a small number of iterations. Thanks to the ensuing acceleration, we can invert an image in as few as 10 optimization steps.

\begin{figure}[ht]
\centering
  \includegraphics[width=\linewidth, trim={5pt 0pt 25pt 0pt}, clip]{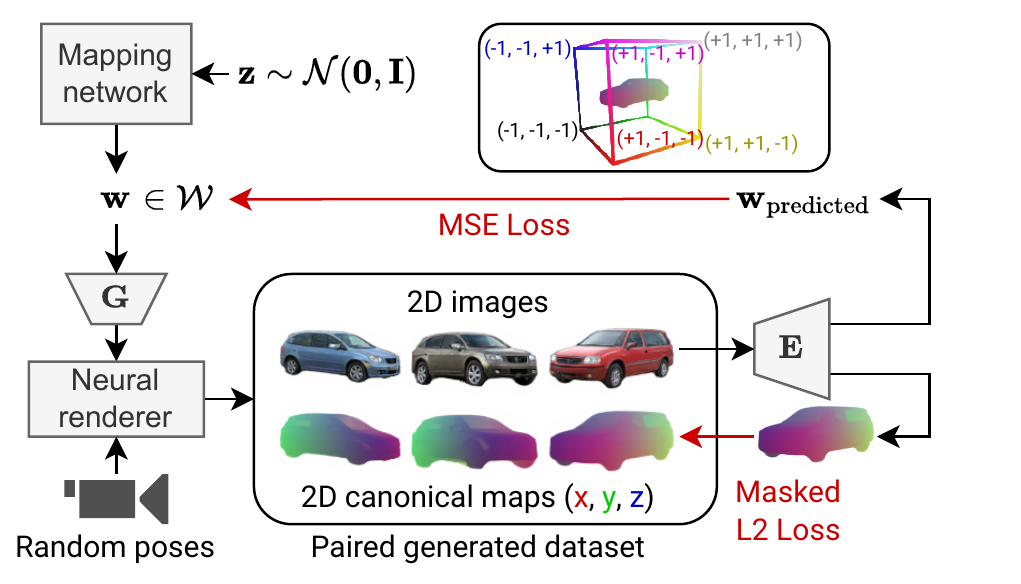}
   \caption{Data generation process for training the encoder ($\mathbf{E}$). We randomly generate synthetic batches of images and associated 2D canonical maps. The encoder is then trained to predict the latent code and canonical map from the RGB image. We then use real images for inference. See also the bounding volume on top, which describes how colors should be interpreted.}
\label{fig:pose-estimation}
\vspace{-2mm}
\end{figure}

\paragraph{Pose estimation.} In previous methods \cite{kanazawa2018cmr, goel2020ucmr, li2020umr, cai2022pix2nerf}, poses are estimated by directly regressing the pose parameters (\eg view matrix or quaternion/scale/translation). While this strategy can learn the task to some extent, it does not effectively incorporate the equivariances required by the problem (\eg translation equivariance) and instead relies on learning them from the data, potentially generalizing poorly in settings other than simple synthetic datasets. More principled approaches can be found in the pose estimation literature, such as NOCS \cite{wang2019nocs}, which frames the problem as a regression of a \emph{canonical map} (NOCS map) in image space, i.e.\ a 2D rendering of the $(x, y, z)$ world-space coordinates of an object (\autoref{fig:pose-estimation}). The mapping is then inverted using a Perspective-\emph{n}-Point (PnP) solver to recover the pose parameters. The main limitation of NOCS \cite{wang2019nocs} is that it requires either ground-truth 3D meshes or hand-modeled synthetic meshes that are representative of the training dataset, since ground-truth canonical maps are not available on real datasets. By contrast, our availability of an object generator allows us to overcome this limitation, as we describe next.

\paragraph{Training and inference.} The main idea underlying our approach -- in contrast to NOCS \cite{wang2019nocs} -- is that we use data generated from our unconditional generator to train the encoder instead of handcrafted data. This allows us to obtain a mapping between latent codes and images, as well as pseudo-ground truth canonical maps that we can use for pose estimation. During training, we sample a minibatch of priors $\mathbf{z} \sim \mathcal{N}(\mathbf{0}, \mathbf{I})$, feed them through the mapping network to obtain the latent codes $\mathbf{w} \in \mathcal{W}$, %
and generate the corresponding RGB images and canonical maps from randomly-sampled viewpoints\footnote{Rendering canonical maps requires only a trivial change to standard NeRF implementations, namely integrating the query coordinates $(x, y, z)$ instead of the RGB channels.}. Finally, we train the network (a SegFormer \cite{xie2021segformer} segmentation network with a custom regression head) to predict the canonical map and the latent code $\mathbf{w}$ from the RGB image. Losses, detailed architecture, and hyperparameters are described in the Appendix \ref{sec:appendix-implementation-details}. For inference, we feed a real image, convert the predicted canonical map to a point cloud, and run a PnP solver to recover all pose parameters (view matrix and focal length).

\subsection{Reconstruction via hybrid GAN inversion}
\label{sec:method-hybrid-inversion}

\begin{figure}[h]
\centering
  \includegraphics[width=\linewidth, trim={0pt 0pt 5pt 0pt}, clip]{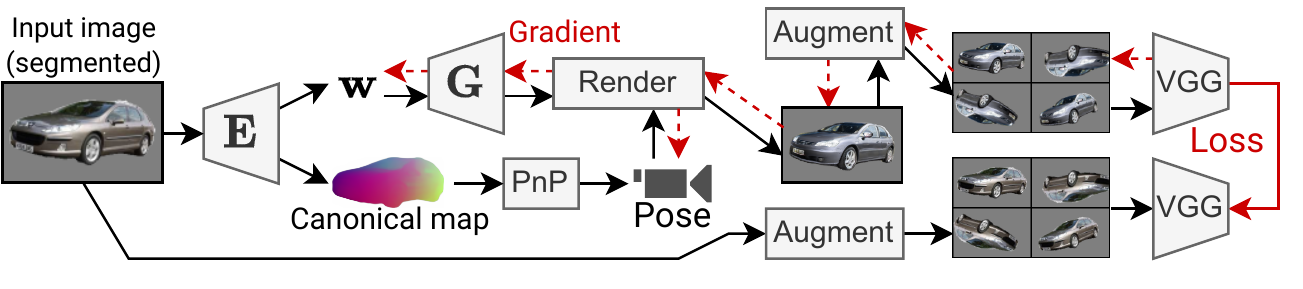}
   \caption{Hybrid inversion process. From the input image, the encoder $\mathbf{E}$ predicts an initial latent code $\mathbf{w}$ and a canonical map, the latter of which is used to recover the pose parameters through a PnP solver. Both $\mathbf{w}$ and the pose are then refined via optimization using a multi-crop VGG loss.}
\label{fig:inversion}
\vspace{-1mm}
\end{figure}

The final step of our pipeline is the refinement of the latent code and pose via gradient-based optimization (\autoref{fig:inversion}). In this step, we found it beneficial to split the initial latent code $\mathbf{w}$ into a different vector for each layer, which we refer to as $\mathbf{w}^+ \in \mathcal{W}^+$. %
For a fixed number of steps $N$, we update $\mathbf{w}^+$ and the pose to minimize a reconstruction error between the rendered image and the input image. We experimented with various loss functions including MSE, L1 and a VGG perceptual loss \cite{zhang2018lpips}, finding that the former two lead to overly blurry results. Eventually, we settled on a VGG loss \cite{zhang2018lpips} with random augmentations, where both the predicted and target images are randomly augmented with geometric image-space transformations (we use 16 augmentations and average their losses). This helps reduce the variance of the gradient, allowing us to further increase the learning rate. We also find that the pose parameterization is an important aspect to consider, and describe it in detail in the Appendix \ref{sec:appendix-implementation-details} (among additional details for this step). 
\section{Experimental setting}
\label{sec:experiments}

We compare against two main directions from the single-view 3D reconstruction literature: real images, following CMR \cite{kanazawa2018cmr}, and synthetic images, following Pix2NeRF \cite{cai2022pix2nerf}.

\paragraph{Real images.} Firstly, we adopt the evaluation methodology of CMR \cite{kanazawa2018cmr} and follow-up works \cite{ goel2020ucmr, li2020umr, bhattad2021viewgen, zhang2022meshinversion}, which focus on real datasets where ground-truth novel views are not available. These methods evaluate the mean IoU between the input images and the reconstructions rendered from the input view. While this metric describes how much the reconstruction matches the input image, it is limited since it does not evaluate how realistic the object looks from other viewpoints. %
Therefore, in the comparison to these works, we also include the FID  \cite{heusel2017ttur} evaluated from random viewpoints, which correlates with the overall generative quality of the reconstructed objects. In this setting, we evaluate our approach on the standard datasets used in prior work -- CUB Birds \cite{wah2011cub} and Pascal3D+ Cars \cite{xiang2014pascal} -- each of which comprise $\sim$5k training images and an official test split which we use for reporting. For the pose distribution used to train the unconditional generator, we rely on the poses estimated by CMR \cite{kanazawa2018cmr} using keypoints. It is worth noting that CMR uses a weak-perspective camera projection model. We found this appropriate for birds, which are often photographed from a large distance, but not for cars, which exhibit varying levels of perspective distortion. Therefore, we upgraded the camera model of P3D Cars to a full-perspective one as described in the Appendix \ref{sec:appendix-implementation-details}.

\paragraph{Extra baselines.} To further demonstrate the applicability of our method to real-world datasets, we establish new baselines on a variety of categories from ImageNet: cars, motorbikes, airplanes, as well as deformable categories such as zebras and elephants. For these classes, we use the splits from \cite{pavllo2021textured3dgan}, which comprise 1-4k training images each (allowing us to assess how well our method fares on small datasets), and use the unsupervised pose estimation technique in \cite{pavllo2021textured3dgan} to obtain the pose distribution, which we also upgrade to a full-perspective camera model. Since no test split is available, we evaluate all metrics on the training split. Moreover, as we observe that the official test set of P3D Cars is too small ($\sim$200 images) to reliably estimate the FID, we construct another larger test set for P3D using non-overlapping images from the car class of ImageNet.

\paragraph{Synthetic images.} Secondly, we evaluate our approach on synthetic datasets: ShapeNet-SRN Cars \& Chairs \cite{shapenet2015, sitzmann2019srn}, and CARLA \cite{dosovitskiy2017carla}. We follow the experimental setting of Pix2NeRF \cite{cai2022pix2nerf}, in which in addition to the FID from random views, pixel-level metrics (PSNR, SSIM) are also evaluated on ground-truth novel views from the test set. On these datasets, we also evaluate the pose estimation performance, as exact ground-truth poses are known. Following \cite{cai2022pix2nerf}, we compute all metrics against a sample of 8k images from the test split, but use all training images. Although ground-truth novel views are available on ShapeNet, we only use such information for evaluation purposes and not for training.

\paragraph{Implementation details.} We describe training hyperparameters as well as additional details in the Appendix \ref{sec:appendix-implementation-details}.
\section{Results}
\label{sec:results}
\vspace{-2mm}
\begin{figure}[ht]
\centering
  \includegraphics[width=\linewidth]{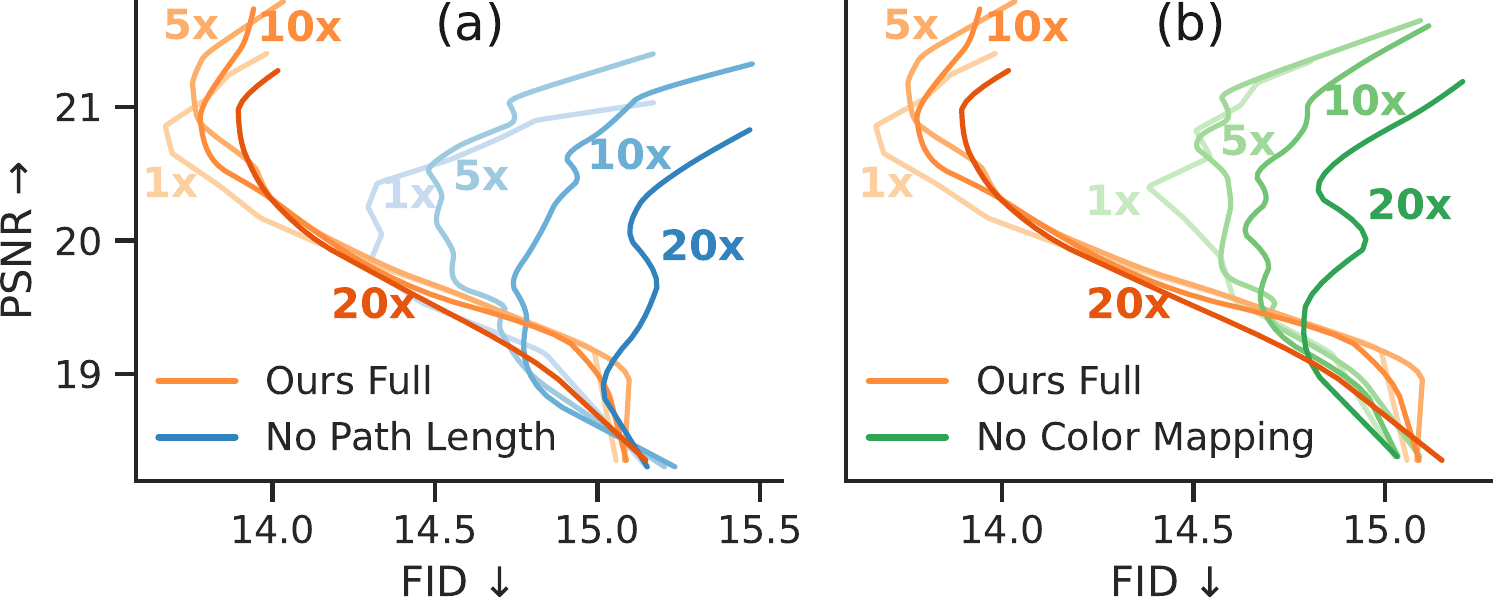}
   \caption{Inversion dynamics and ablations on P3D Cars on a larger test set from ImageNet, under different learning rate gains (1x, 5x, 10x, 20x) for the latent code $\mathbf{w}$. All curves start from the bottom-right corner. When path length regularization is applied (\textbf{a}), the curves exhibit a higher linearity, which allows us to increase the learning rate while reducing the number of optimization steps. Conversely, when the regularizer is not adopted, the curves are more spaced apart and performance degrades quickly as the gain increases. Furthermore, our color mapping module (\textbf{b}) allows for a better reconstruction. We also identify an overfitting region, where the PSNR keeps increasing but the FID starts degrading, indicating that there is a trade-off between these metrics.}
\label{fig:ablation-ppl-color}
\vspace{-1.5mm}
\end{figure}

\begin{figure*}[t]
\centering
  \includegraphics[width=\linewidth, trim={0pt 4pt 0pt 4pt}, clip]{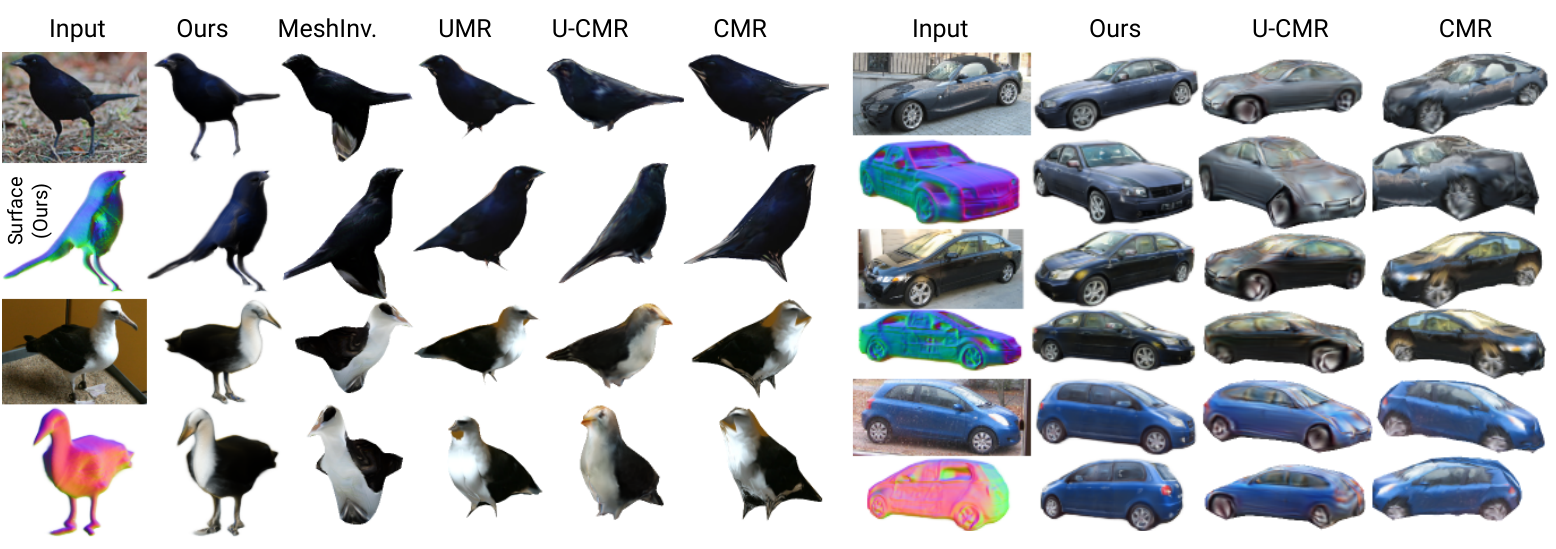}
   \caption{Qualitative results and side-by-side comparison on the test set of CUB (\textbf{left}) and Pascal3D+ Cars (\textbf{right}), at 128$\times$128. The first row of each sample is rendered from the input viewpoint, whereas the second row illustrates a random view. Compared to the other works, which adopt a triangle mesh representation with a fixed topology, our SDF parameterization can model arbitrary topologies and can easily represent fine details such as the legs of the birds or the geometry of the cars, without enforcing any symmetry constraints. We observe occasional artifacts in the surface that are not visible from the RGB image, \eg concave areas in the wings of birds or near the headlights of cars, which arise from the unconditional generator and can in principle improve with better supervision techniques.}
\label{fig:qualitative-results-p3d-cub}
\vspace{-1.5mm} %
\end{figure*}

\paragraph{Inversion dynamics and settings.} Before presenting our main results, we carry out a preliminary study on how to achieve the best speed on the hybrid inversion task. In \autoref{fig:ablation-ppl-color}, we analyze the inversion dynamics under different gain factors for the learning rate of the latent code $\mathbf{w}$ (1x, 5x, 10x, 20x) along with a corresponding reduction in the number of optimization steps. When both path length regularization and color mapping are used, we find the dynamics to be almost linear up to a certain point. Both the FID (evaluated on random views) and PSNR (computed on the input view) improve monotonically, eventually reaching a ``sweet spot'' after which the FID starts degrading, indicating overfitting. When we remove these components, the inversion dynamics become less predictable and the overall performance is affected when higher gains are used. We also find that using a lower learning rate is generally better, but requires more iterations. As a result, we propose the following settings: a higher-quality but slower schedule, \emph{Hybrid Slow}, with $N$=30 inversion steps at 5x gain, and \emph{Hybrid Fast}, where we ramp up the gain to 20x and use only $N$=10 steps. We also experimented with higher gains (up to 50x), but could not get these to reliably converge. Furthermore, for a fair comparison with works that are purely feed-forward-based, we also report a baseline with $N$=0, \ie we evaluate the output of the encoder with no inversion.

\paragraph{Quantitative evaluation (real images).} \autoref{tab:evaluation-real} (top) shows our main comparison on datasets of real images, following the CMR \cite{kanazawa2018cmr} protocol. On P3D Cars and CUB, our initial guess of the pose and latent code ($N$=0) already provides an advantage over existing approaches, with a 36\% decrease in FID on CUB over the state-of-the-art, and a 9\% increase in IoU on P3D Cars, despite our model not being trained to optimize the latter metric (unlike the other approaches, which are all encoder-based and include a supervised loss). %
We attribute this improvement to our more powerful NeRF-based representation (as opposed to sphere-topology triangle meshes used in prior works), as well as a better pose estimation performance. Following refinement via hybrid inversion, performance is further boosted in as few as 10 steps. Finally, we also establish new baselines on categories from ImageNet (\autoref{tab:evaluation-real}, bottom), demonstrating that our method is effective beyond benchmark datasets.

\begin{table}[t]
\centering
\resizebox{\linewidth}{!}{
\begin{tabular}{l|ll|ll}
        & \multicolumn{2}{c|}{Pascal3D+ Cars} & \multicolumn{2}{c}{CUB Birds} \\
Method  & IoU $\uparrow$         & FID $\downarrow$       & IoU $\uparrow$              & FID $\downarrow$               \\ \hline\hline
CMR \cite{kanazawa2018cmr}     & 0.64        & 273.28              & 0.706             & 105.04            \\
U-CMR \cite{goel2020ucmr}   & 0.646       & \underline{223.12}             & 0.644             & 69.42             \\
UMR \cite{li2020umr}     & -           & -                   & \underline{0.734}             & \underline{43.83}             \\
SDF-SRN \cite{lin2020sdfsrn} & \underline{0.81} & 254.90 & - & - \\
ViewGeneralization \cite{bhattad2021viewgen} & 0.78        & -                   & 0.629             & -                 \\
StyleGANRender \cite{zhang2021imagegan} & 0.80 & - & - & - \\ \hline
Ours Init.\ ($N$=0) & \textbf{0.883}        & \textbf{75.90} (15.08)             & \textbf{0.739}             & \textbf{28.15} \\ \hline\hline
MeshInv. ($N$=200) ($\ast$) ($\dag$) \cite{zhang2022meshinversion} & -        & -           & 0.752             & 31.60 \\  \hline
Ours Hybrid Slow ($N$=30) ($\dag$)  & \textbf{0.920}        & \underline{73.53} (14.36)          & \textbf{0.844}             & \textbf{24.70}                 \\
Ours Hybrid Fast ($N$=10) ($\dag$)  & \underline{0.917}        & \textbf{73.12} (14.36)      & \underline{0.835}             & \underline{25.65}                 \\
\end{tabular}
}
\resizebox{\linewidth}{!}{
\setlength\tabcolsep{3pt}
\begin{tabular}{l|ll|ll|ll|ll|ll}
        & \multicolumn{2}{c|}{Car} & \multicolumn{2}{c|}{Motorcycle} & \multicolumn{2}{c|}{Airplane} & \multicolumn{2}{c|}{Zebra} & \multicolumn{2}{c}{Elephant} \\
Method  & IoU $\uparrow$         & FID $\downarrow$       & IoU $\uparrow$              & FID $\downarrow$      & IoU $\uparrow$         & FID $\downarrow$ & IoU $\uparrow$         & FID $\downarrow$ & IoU $\uparrow$         & FID $\downarrow$         \\ \hline\hline
Init.\ $N$=0 & 0.933 &  9.88 & 0.804 &	40.65 & 0.749 &	\textbf{18.77} & 0.724 &	\textbf{21.58} & 0.781 &	107.34   \\
Slow $N$=30 ($\dag$)  & \textbf{0.953} &	\textbf{8.77} & \textbf{0.851} &	\textbf{38.6} & \textbf{0.813} & \underline{19.78} & \textbf{0.802} &	\underline{24.47} & \textbf{0.848} &	\textbf{99.77} \\
Fast $N$=10 ($\dag$) &  \underline{0.952} &	\underline{8.91} & \underline{0.85} &	\underline{39.72} & \underline{0.805} &	21.33 & \underline{0.793} &	26.41 & \underline{0.845} &	\underline{104.12} \\
\end{tabular}
}
\caption{Evaluation on real datasets (CMR setting with predicted camera) on P3D/CUB (upper table) and ImageNet (bottom table). The first rows are purely feed-forward-based, while the remaining are inversion-based. All FIDs have been computed by us at 128$\times$128 under uniform settings, wherever a public implementation was available. Note that the seemingly high FIDs on P3D are due to the small size of the test set ($\sim$200 images), and therefore in parentheses we report an additional FID evaluated against a non-overlapping test set from ImageNet Cars. \textbf{Legend:} ($\ast$) Uses class-conditional model; ($\dag$) Uses optimization for $N$ iterations.} %
\label{tab:evaluation-real}
\vspace{-1.5mm}
\end{table}

\begin{figure}[t]
\centering
  \includegraphics[width=\linewidth, trim={0pt 3pt 0pt 3pt}, clip]{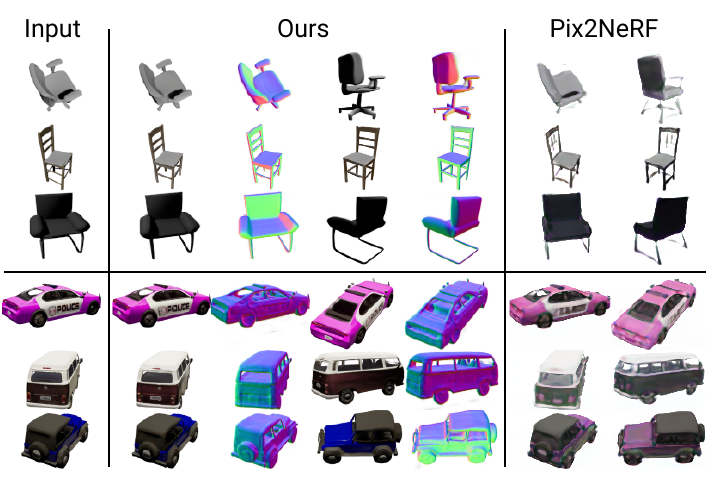}
  \vspace{-6mm}
   \caption{Qualitative results on synthetic datasets (test set of ShapeNet Chairs \& CARLA) and side-by-side comparison to Pix2NeRF \cite{cai2022pix2nerf} on input and random views at 128$\times$128. We observe that our method better predicts fine details such as the legs of the chairs, the text on cars, and color distributions.}
\label{fig:qualitative-results-synthetic}
\vspace{1mm}
\resizebox{\linewidth}{!}{
\setlength\tabcolsep{4pt}
\begin{tabular}{l|lll|lll|l}
        & \multicolumn{3}{c|}{SRN Cars} & \multicolumn{3}{c|}{SRN Chairs} & \multicolumn{1}{c}{CARLA} \\
Method  & PSNR $\uparrow$ & SSIM $\uparrow$ & FID $\downarrow$         & PSNR $\uparrow$ & SSIM $\uparrow$ & FID $\downarrow$ & FID $\downarrow$  \\ \hline\hline
Pix2NeRF \cite{cai2022pix2nerf}     & - & - & - & 18.14 &	0.84 &	26.81 & 38.51    \\ \hline
Ours Init.\ ($N$=0) & 18.54 &	0.848 &	12.39      & 18.26	& 0.857 &	8.64 & 6.49 \\ 
Ours Hybrid Slow ($N$=30)  & \textbf{19.55} &	\textbf{0.864} &	\textbf{11.37}     & \textbf{19.36} &	\textbf{0.875}	& \textbf{7.44} & \textbf{5.97}                 \\
Ours Hybrid Fast ($N$=10)  & \underline{19.24} &	\underline{0.861} &	\underline{12.26}     & \underline{19.02} &	\underline{0.871} &	\underline{7.62} & \underline{6.18}                 \\
\end{tabular}
}
\vspace{-2.5mm}
\captionsetup{type=table}\caption{Evaluation on synthetic datasets. All metrics are computed at 128$\times$128 using \emph{predicted} poses. PSNR and SSIM are evaluated on novel views (not available on CARLA), and the FID on random views. Since \cite{cai2022pix2nerf} is not evaluated on SRN Cars, we establish baselines on this category.}
\label{tab:evaluation-synthetic}
\vspace{-3.5mm}
\end{figure}

\paragraph{Quantitative evaluation (synthetic images).} In \autoref{tab:evaluation-synthetic}, we further evaluate our approach against \cite{cai2022pix2nerf} on synthetic data. Again, even before applying hybrid inversion, we observe an improvement in the FID (-68\% on chairs and -83\% on CARLA) as well as in the novel-view evaluation (PSNR, SSIM). Applying hybrid inversion further widens this gap.

\begin{figure}[t]
\centering
  \includegraphics[width=\linewidth]{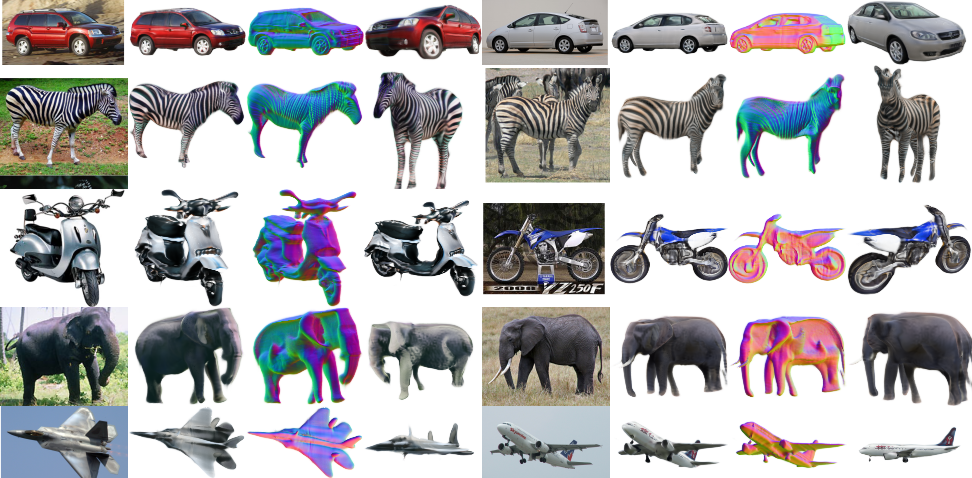}
  \vspace{-5.5mm}
   \caption{Additional qualitative results produced by our method on ImageNet.  More results can be found in the appendix. Most classes learn a correct geometry despite being trained with only 1-4k images. We only observe some spurious concavities in the shape of the elephants, as well as a failure to correctly disentangle the stripes of the zebra from the surface.}
\label{fig:qualitative-results-additional}
\vspace{1mm}
\resizebox{\linewidth}{!}{
\begin{tabular}{l|lll}
        Method & SRN Cars $\downarrow$ & SRN Chairs $\downarrow$ & CARLA $\downarrow$  \\ \hline\hline
Pix2NeRF Encoder \cite{cai2022pix2nerf} & - & \underline{15.55}$^\circ$ & 4.23$^\circ$    \\ \hline
Direct pose regression & \underline{17.08}$^\circ$        & 19.51$^\circ$           &     \underline{3.21}$^\circ$   \\ 
Ours NOCS + PnP  & \textbf{10.84}$^\circ$        & \textbf{7.29}$^\circ$           &     \textbf{1.08}$^\circ$     \\
\end{tabular}
}
\vspace{-2.5mm}
\captionsetup{type=table}\caption{Pose estimation accuracy (mean rotation error in degrees) on synthetic datasets, where ground-truth poses are available. All methods are feed-forward (no inversion). Results for \cite{cai2022pix2nerf} are computed after a rigid alignment to the ground-truth reference frame.} 
\label{tab:evaluation-pose}
\vspace{-4.5mm}
\end{figure}

\paragraph{Qualitative results.} \autoref{fig:qualitative-results-p3d-cub} shows a side-by-side comparison to \cite{kanazawa2018cmr, goel2020ucmr, li2020umr, zhang2022meshinversion} on P3D/CUB, while \autoref{fig:qualitative-results-synthetic} shows a comparison to \cite{cai2022pix2nerf} on synthetic datasets. To further demonstrate the applicability of our approach to real-world images, in \autoref{fig:qualitative-results-additional} we display extra results on ImageNet. Furthermore, for our method, we also show the surface normals obtained by analytically differentiating the SDF.
We refer the reader to the respective figures for a discussion of the advantages and shortcomings of our method. Finally, we include additional qualitative results in Appendix \ref{sec:appendix-sub-qualitative}.

\paragraph{Pose estimation.} We evaluate pose estimation in \autoref{tab:evaluation-pose}. For this experiment, we use synthetic datasets for which exact ground-truth poses are known. We compare our NOCS-inspired approach to two baselines: \emph{(i)} direct regression of pose parameters (using a quaternion-based parameterization, see Appendix \ref{sec:appendix-implementation-details}), where we keep the SegFormer backbone unchanged and only switch the output regression head for a fair comparison, and \emph{(ii)} Pix2NeRF's encoder \cite{cai2022pix2nerf}, which is trained to predict azimuth/elevation, a less expressive pose representation specific to the pose distribution of these datasets. %
We evaluate the mean rotation angle between predicted and ground-truth orientations, and observe that our NOCS-inspired approach achieves a significantly better error (53\% and 74\% reduction on chairs and CARLA, respectively) while being more general. Interestingly, our direct pose regression baseline achieves a similar performance to Pix2NeRF's encoder despite using a more expressive transformer architecture, suggesting that the main bottleneck lies in the pose representation itself and not in the architecture. As a side note, we also observe that the NOCS-based model converges much faster than the pose regression baseline, as the NOCS framework better incorporates equivariances to certain geometric transformations, while the baseline method has to learn them from the data. %

\paragraph{Ablations.} In addition to those in \autoref{fig:ablation-ppl-color}, we conduct further ablation experiments in Appendix \ref{sec:appendix-sub-ablations}. Among other things, we evaluate the impact of SDFs, compare our hybrid inversion method to an encoder baseline with a comparable architecture, and assess the impact of pose estimation.

\paragraph{Conversion to triangle mesh.} We can easily convert our reconstructions to triangle meshes in a principled way by extracting the 0-level set of the SDF and using marching cubes \cite{lorensen1987marching}, as we show in the Appendix \ref{sec:appendix-sub-qualitative}.

\paragraph{Failure cases.} We show and categorize these in Appendix \ref{sec:appendix-sub-failure-cases}. Furthermore, to guide future research, in Appendix \ref{sec:appendix-negative-results} we discuss ideas that we explored but did not work out.

\section{Conclusion}
\label{sec:conclusion}

We introduced a framework for reconstructing shape, appearance, and pose from a single view of an object. Our approach leverages recent advances in NeRF representations and frames the problem as a 3D-aware GAN inversion task. In a hybrid fashion, we accelerate this process by learning an encoder that provides a first guess of the solution and incorporates a principled pose estimation technique. We achieve state-of-the-art performance on both synthetic and real benchmarks, and show that our approach is efficient (requiring as few as 10 inversion steps to reconstruct an image) and effective on small datasets. In the future, we would like to scale to higher resolutions and improve the reconstructed surface quality, \eg by leveraging semi-supervision on extra views or shape priors. We would also like to explore ways to automatically infer the pose distribution from the data.

{\small
\bibliographystyle{ieee_fullname}
\bibliography{egbib}
}

\clearpage
\newpage
\appendix
\section{Supplementary material}
\label{sec:appendix}

\subsection{Implementation details}
\label{sec:appendix-implementation-details}

\paragraph{Dataset preparation.}  For CUB \cite{wah2011cub}, we use the segmentation masks and poses from CMR \cite{kanazawa2018cmr} estimated using structure-from-motion. These poses adopt a weak-perspective camera model, which we keep as-is (our neural renderer implementation supports multiple projection models). Similarly, for P3D Cars \cite{xiang2014pascal}, we use poses from CMR but obtain the segmentation masks using Mask R-CNN \cite{he2017maskrcnn} as was done in previous work. Additionally, we upgrade its camera projection model to a full perspective model by freezing the rotations and re-estimating all the other parameters using the procedure described in the next paragraph. For ImageNet, we estimate poses from scratch using the same procedure and predict segmentation masks using PointRend \cite{kirillov2020pointrend}. Additionally, we augment all real datasets mentioned so far with horizontal flips. We do not augment synthetic datasets (CARLA \cite{dosovitskiy2017carla} and ShapeNet \cite{shapenet2015}).

\paragraph{Pose estimation and parameterization.} While most neural renderers adopt \emph{camera-to-world} view matrices (plus focal length), this representation is not necessarily the best for optimization purposes. Among various issues, we mention the necessity to enforce orthogonality constraints in the rotation matrix, a dependency between rotation and translation (which can be solved by switching to a \emph{world-to-camera} representation), and an entanglement between translation and focal length (ideally, as depth increases, the learning rate for the translation needs to be amplified in order to keep a linear behavior in projective space). Therefore, for all our steps where pose optimization is involved (namely, the initial data preparation and the hybrid inversion step), we adopt a custom pose parameterization that tackles these issues and is easier to optimize, while being fully differentiable and convertible to view matrices for use in neural renderers. We also make sure that our neural renderer is fully differentiable w.r.t.\ the pose, even when coarse-fine importance sampling is used and view-dependent effects are enabled, as we found that existing implementations present gradient detachments in some nodes of the pipeline.

Our pose representation can be regarded as an augmentation of a weak-perspective camera model, and describes a \emph{world-to-camera} transformation parameterized by a rotation $\mathbf{q} \in \mathbb{R}^4$ (a unit quaternion), a screen-space scale $s \in \mathbb{R}$, a screen-space translation $\mathbf{t}_2 \in \mathbb{R}^2$, and a perspective distortion factor $z_0 \in \mathbb{R}$. At runtime, we derive the focal length $f = 1 + \exp(z_0)$ and the 3D translation $\mathbf{t}_3 = [\mathbf{t}_2/s; f/s]$. Such a parameterization is equivalent to a full-perspective model, but results in more ``linear'' optimization dynamics.

For the datasets where we estimate the poses ourselves (ImageNet and, partially, P3D Cars), we use the template-based pose estimation technique described in \cite{pavllo2021textured3dgan} with our pose parameterization.

\begin{figure}[t]
\centering
  \includegraphics[width=0.8\linewidth]{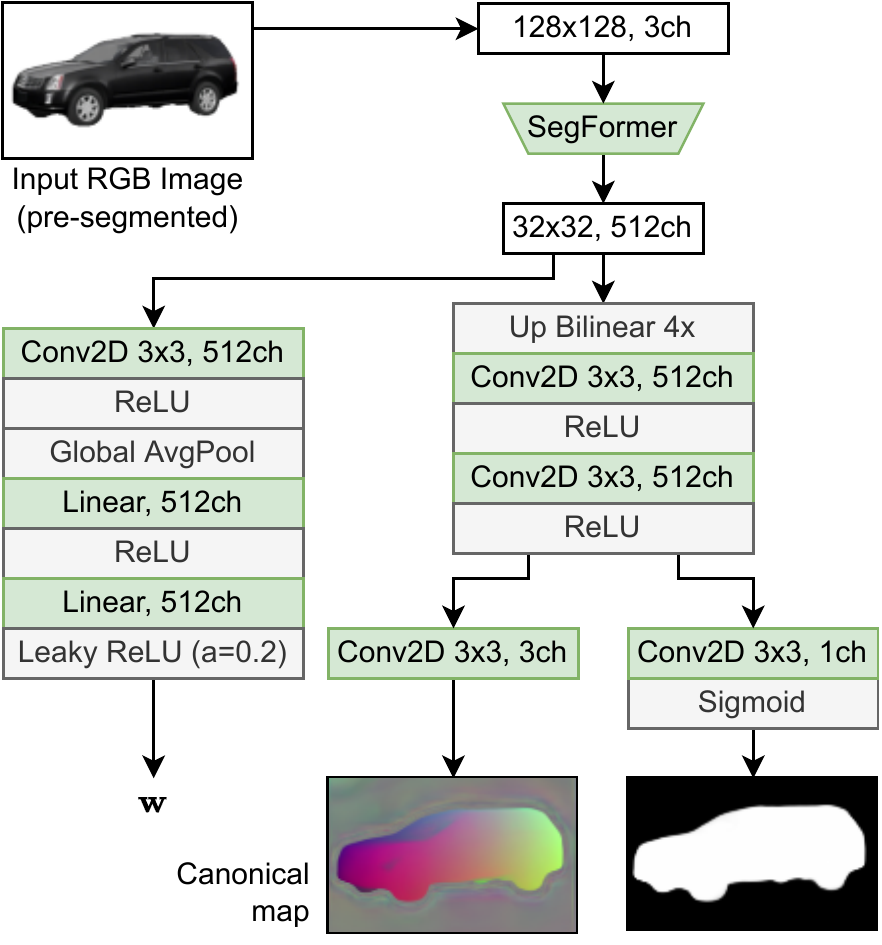}
   \caption{Architecture of the encoder used for bootstrapping the latent code and pose estimation. Note that we include a Leaky ReLU activation in the final layer of the latent code regressor, which mimics the behavior of the mapping network.}
\label{fig:appendix-encoder}
\end{figure}

\begin{figure*}[t]
\centering
  \includegraphics[width=\linewidth]{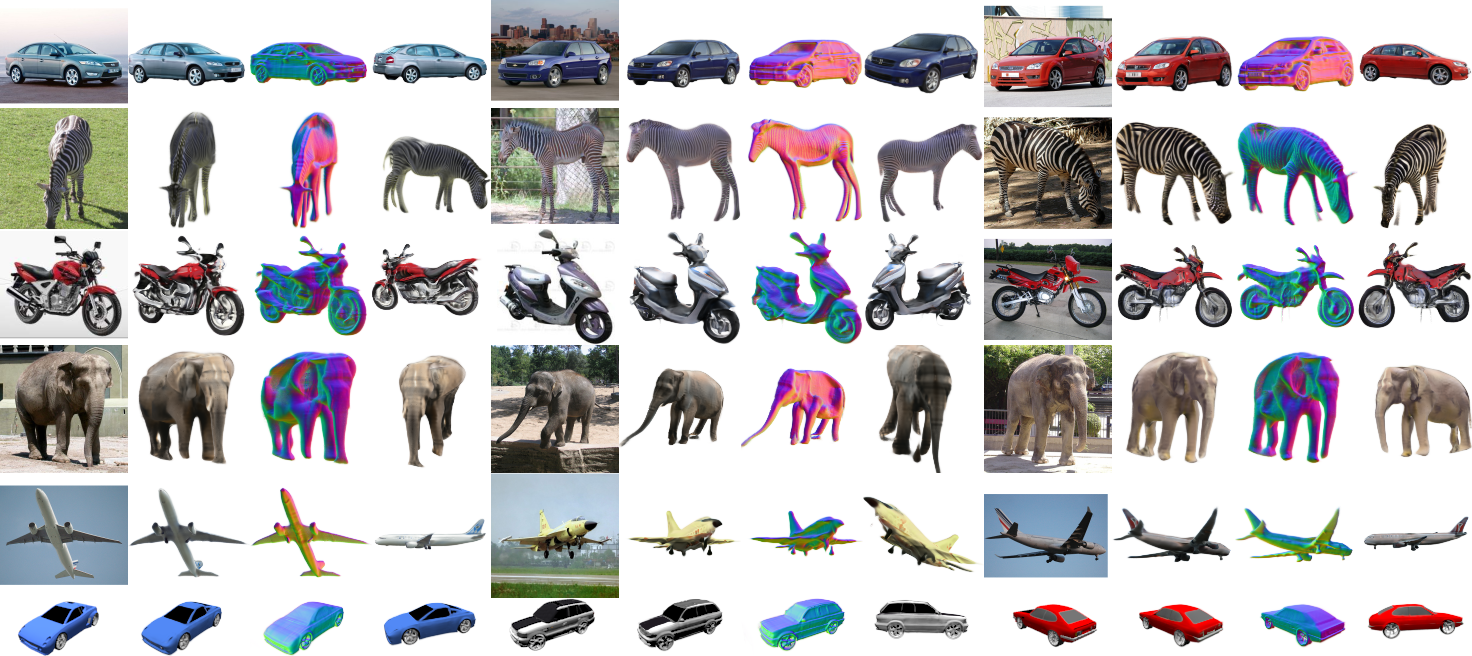}
   \caption{Additional qualitative results produced by our method on ImageNet (top rows) and ShapeNet Cars (last row).}
\label{fig:appendix-imagenet}
\end{figure*}

\paragraph{Unconditional generator.} We train the unconditional generator for 300k iterations, except for CUB and ImageNet elephants, where we use 200k. Similarly, we adopt R1 regularization on all datasets with $\gamma = 5$, except on elephants where we use $\gamma = 10$. Optimizer, learning rate, and batch size are the same as in \cite{chan2022eg3d}. We found it beneficial for stability to warm up the learning rate of both the generator and discriminator, starting from 1/10th of the specified value and linearly increasing it over 2000 iterations.

As in \cite{chan2022eg3d}, we condition the discriminator on the pose, but we nonetheless observe that all our models converge even without such conditioning, although this sometimes leads to surface artifacts (such as objects appearing concave).
On all datasets except ShapeNet, we enable adaptive discriminator augmentation (ADA) \cite{karras2020ada}, which reduces discriminator overfitting by enhancing its input images with differentiable augmentations. We only adopt geometric transformations (scale, translation, rotation). However, we observe that the implementation of ADA in \cite{chan2022eg3d} is not 3D-aware, as the augmentations are only carried out in image space, while the discriminator is conditioned on the original camera pose, leading to artifacts. We implemented a 3D-aware version of ADA where both the image and camera pose are augmented with the same transformation, and the discriminator is conditioned on the augmented pose.

\begin{figure}[b]
\centering
  \includegraphics[width=0.49\linewidth]{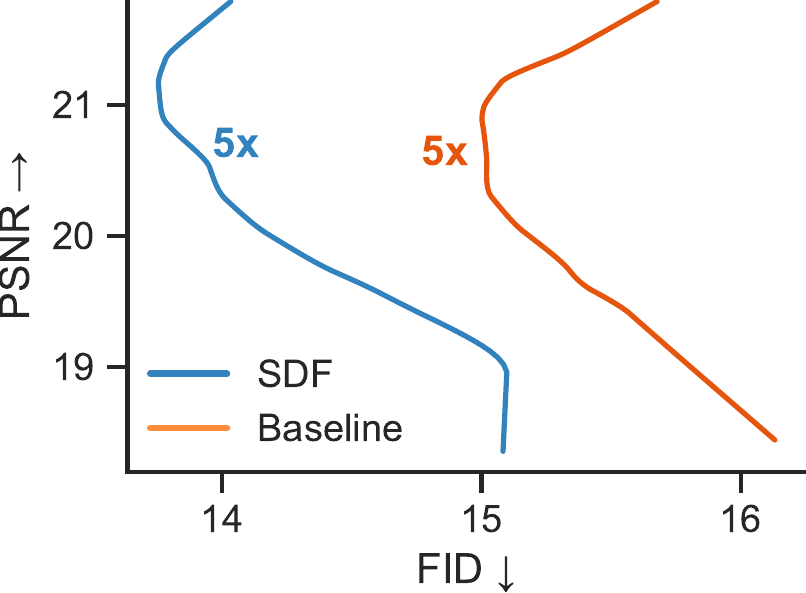}
  \includegraphics[width=0.49\linewidth]{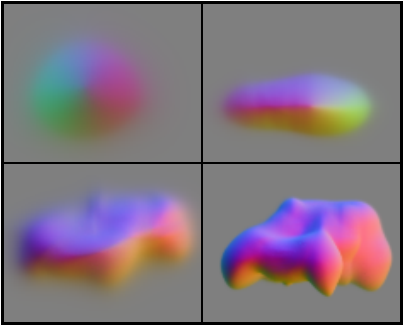}
   \caption{\textbf{Left:} impact of the SDF representation on the inversion dynamics (at gain 5x). As in \autoref{fig:ablation-ppl-color}, the analysis is carried out on our larger P3D Cars test set. \textbf{Right:} spherical initialization of the SDF and its evolution as training progresses.}
\label{fig:appendix-sdf}
\end{figure}

\paragraph{SDF details.} As for the SDF representation \cite{yariv2021volsdf}, the volume density is modulated by two learnable parameters $\alpha, \beta > 0$. Unlike \cite{yariv2021volsdf}, which ties $\alpha = \beta$ (according to \autoref{eq:sdf}), we found it helpful for convergence to learn them separately. We initialize $\alpha = 1$ and $\beta = 0.1$, and clamp their lower bound to $10^{-3}$ for stability during training. Before training the unconditional model, we initialize the SDF to a unit sphere through optimization. We pre-train the model for 1000 iterations using the following loss:
\begin{equation}
    \mathcal{L}_\text{SDF} = \mathbb{E}_\mathbf{x}\bigl[\bigl(d(\mathbf{x}) - (\lVert \mathbf{x} \rVert - 1)\bigr)^2\bigr].
\end{equation}
A visualization of the SDF can be seen in \autoref{fig:appendix-sdf} (right). For the Eikonal loss, we use a weight of 0.1 as recommended by \cite{yariv2021volsdf}. 

\paragraph{View-dependent effects.} We also incorporate the ability to efficiently model view-dependent effects. The view direction of each pixel (which is constant across depth, and can therefore be computed only once per pixel) is processed through a small feed-forward network to produce a 32-dimensional vector, and summed with another 32-dimensional vector coming from the triplanar decoder. The result is then processed through a Leaky ReLU activation and another linear layer to produce the final output. This late fusion strategy ensures that memory consumption is minimal. We enable this feature only on CARLA, as ShapeNet does not have any specular reflections and the other datasets are too small to properly disentangle appearance and specular effects.

\paragraph{Bootstrapping and pose estimation.} For the encoder architecture, we adopt SegFormer B5 \cite{xie2021segformer}, a recently-proposed transformer-based backbone for semantic segmentation. The output feature map from the backbone is connected to two heads: a fully-connected one that regresses the latent code $\mathbf{w}$ and a convolutional one that regresses the canonical map and the associated segmentation mask. The detailed architecture is shown in \autoref{fig:appendix-encoder}. We initialize the backbone using ImageNet weights and train the model end-to-end for 120k iterations with a batch size of 32 samples, using Adam optimizer. We adopt an initial learning rate of 6e-5, which we decay to 6e-6 after 60k iterations. As for the losses, we use a simple mean squared error (MSE) loss for the latent code ($\mathcal{L}_\text{latent}$), an L1 loss for the segmentation mask ($\mathcal{L}_\text{mask}$), and a masked L2 loss (with square root) for the canonical map ($\mathcal{L}_\text{map}$), \ie a rotation-invariant version of the L1 loss, for better robustness to artifacts coming from the generator:
\begin{align}
    \mathcal{L}_\text{latent} &= \lVert \mathbf{\hat{w}} - \mathbf{w} \rVert^2 , \\
    \mathcal{L}_\text{mask} &= \frac{1}{W H} \sum_{i=1}^W\sum_{j=1}^H | \hat{m}_{i, j} - m_{i, j} | , \\
    \mathcal{L}_\text{map} &= \frac{1}{W H} \sum_{i=1}^W\sum_{j=1}^H m_{i, j} \lVert \mathbf{\hat{p}}_{i, j} - \mathbf{p}_{i, j} \rVert , \\
    \mathcal{L}_\text{total} &= \mathcal{L}_\text{latent} + \mathcal{L}_\text{mask} + \mathcal{L}_\text{map}
\end{align}
where $\mathbf{\hat{w}}$ is the predicted latent code, $\mathbf{\hat{p}}_{i, j}$ is the predicted canonical map at the $i, j$ image-space coordinates (a 3D vector for each position), $\mathbf{p}$ is the ground-truth one, $\hat{m}$ is the predicted mask, and $m$ is the ground-truth mask.
This contrasts NOCS \cite{wang2019nocs}, which frames the task as a classification problem using quantized coordinates. For inference, the regressed canonical map is thresholded using the predicted mask, converted to a dense point cloud, and used as the input for SQPnP \cite{terzakis2020sqpnp}, a fast PnP solver that retrieves a global optimum (we use the implementation in OpenCV \cite{opencv_library}). Since SQPnP -- as most PnP methods -- requires the focal length to be pre-determined, we select 10 representative focal lengths from the training set (one for each 10th percentile), run the algorithm for each, and select the solution with the lowest reprojection error.

\paragraph{GAN inversion.} For the hybrid inversion step, we use Adam optimizer \cite{kingma2014adam} with a base learning rate of 0.02, and optionally amplify the learning rate of the latent code $\mathbf{w}$ by a gain factor (as reported in the individual experiments). We additionally set $\beta_2 = 0.95$ for a faster reaction. We do not dynamically adjust the hyperparameters throughout the procedure. For the pose, we use our previously-described parameterization as this results in better optimization dynamics. We optimize the following objective:
\begin{equation}
    \min_{\mathbf{w}, \mathbf{q}, s, \mathbf{t}_2, z_0}\; \frac{1}{K} \sum_{k=1}^{K}\, \text{LPIPS}(\mathbf{c}_\text{pred}^{[k]}, \mathbf{c}_\text{gt}^{[k]})\, ,
\end{equation}
where $\mathbf{c}_\text{pred}$ represents the predicted image, $\mathbf{c}_\text{gt}$ is the ground-truth one, $k$ is the augmentation index (we use $K = 16$ augmentations), and the LPIPS operator \cite{zhang2018lpips} describes the distance between the VGG embeddings of the two images.
After each iteration, we reproject the pose parameters onto the valid set of constraints ($\mathbf{q}$ unit length).

\paragraph{Evaluation details.} For the evaluation on real datasets (P3D, CUB, ImageNet), we follow the protocol of \cite{pavllo2020convmesh, pavllo2021textured3dgan} and evaluate the FID on an empty background (value $=0$, \ie gray). All scores are evaluated at $128\times128$, using antialiased resampling (also referred to as \emph{area} interpolation) if resizing is needed, as the FID is sensitive to this aspect. For the approaches we compare to, we compute their FID under the same settings by modifying their public implementations.

\begin{figure}[b]
\centering
  \includegraphics[width=0.49\linewidth]{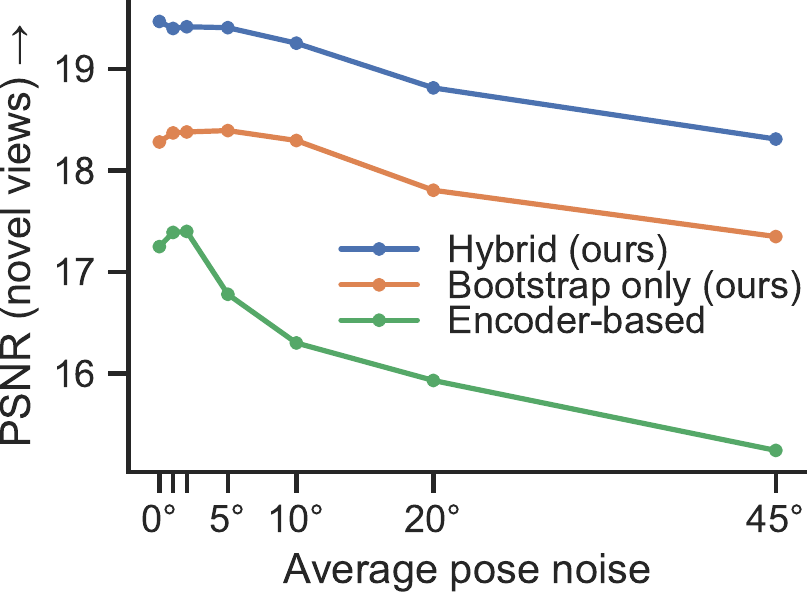}
  \includegraphics[width=0.49\linewidth]{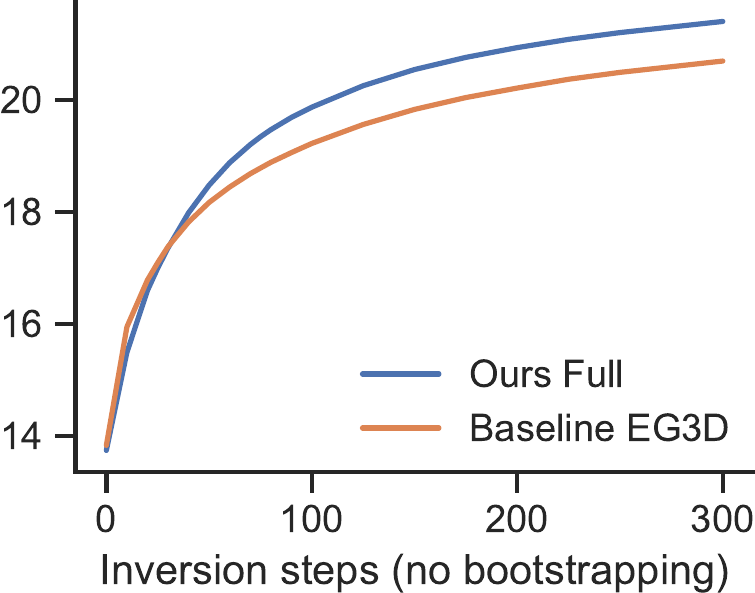}
   \caption{Additional ablations on ShapeNet Chairs, where we evaluate the PSNR on novel views from the test set. \textbf{Left:} comparison of our hybrid inversion approach (and initial bootstrapping without refinement) to an encoder-based baseline under simulated pose perturbations. \textbf{Right:} inversion on a vanilla EG3D backbone \emph{vs} our proposed architecture. Since the goal of this experiment is to evaluate only the impact of the unconditional generator, we start from an average latent code (\ie no bootstrapping) and use the ground-truth pose.}
\label{fig:appendix-ablations}
\end{figure}

\subsection{Additional results}
\label{sec:appendix-additional-results}

\subsubsection{Ablation experiments}
\label{sec:appendix-sub-ablations}

\paragraph{Encoder-based architecture.} For our next experiment, we build an encoder-based variant of our architecture and switch to a conditional GAN setting. Instead of using a mapping network to map $\mathbf{z}$ to $\mathbf{w}$, we learn a convolutional encoder that takes a 2D image as input and directly predicts $\mathbf{w}$. We also experiment with various supervision strategies, including a dual discriminator (an unconditional one that discriminates random views plus a conditional one that discriminates the input view), and a single unconditional discriminator with an L1 or MSE loss to fit the input image, as in Pix2NeRF \cite{cai2022pix2nerf}. Based on early experiments, we found that the dual discriminator approach yields the best results (as it does not require balancing the losses), and we use this strategy throughout our ablations. Moreover, for a fair comparison, we use the same backbone for the conditional (encoder-based) and unconditional (inversion-based) experiments.\vfill

\paragraph{Encoder- \emph{vs} inversion-based baselines.} In \autoref{fig:appendix-ablations} (left), we compare our hybrid inversion approach to the aforementioned encoder-based baseline. We conduct this experiment on ShapeNet Chairs, where exact ground-truth poses are known. In this setting, we randomly perturb individual poses by injecting noise at different levels (from 0$^\circ$ to 45$^\circ$) without altering the overall pose distribution, and study which approach is more robust to inaccurate poses as noise increases. Importantly, for a fair comparison to encoder methods (which are feed-forward), we also include a \emph{bootstrap only} baseline where we do not refine the initial guess of our solution.  We immediately observe that our \emph{bootstrap only} baseline achieves a higher PSNR compared to the encoder-based approach. Furthermore, the performance of our approach decreases gracefully as noise increases, whereas with the encoder-based method we can observe a sharp degradation as early as 5$^\circ$. Based on these findings, we conclude that inversion-based approaches are more appropriate for real datasets where poses are potentially inaccurate, as these methods rely on the overall pose distribution as opposed to the correctness of individual poses.

\paragraph{Impact of the backbone.} In \autoref{fig:appendix-ablations} (right), we assess the impact of the backbone on the final reconstruction result. We leave out some of our contributions (bootstrapping, pose estimation, and hybrid inversion) and purely focus on our proposed backbone components (SDF, color mapping, optimized path length regularization, as well as the minor improvements over \cite{chan2022eg3d}). To this end, we conduct a vanilla inversion experiments on ShapeNet Chairs, using ground-truth poses and starting from the ``average'' latent code in $\mathcal{W}$. We compare our full backbone to a vanilla EG3D, and observe an advantage when adopting our proposed changes.

\paragraph{SDF representation.} Similar to \autoref{fig:ablation-ppl-color}, we conduct an analysis of the inversion dynamics with and without our proposed SDF representation (\autoref{fig:appendix-sdf}). We find that the optimization dynamics are similar, but the SDF baseline gets an FID boost owing to a better unconditional generator, in addition to the other practical benefits (\eg ability to easily extract surface, normals, and mesh).

\begin{figure}[t]
\centering
  \includegraphics[width=\linewidth]{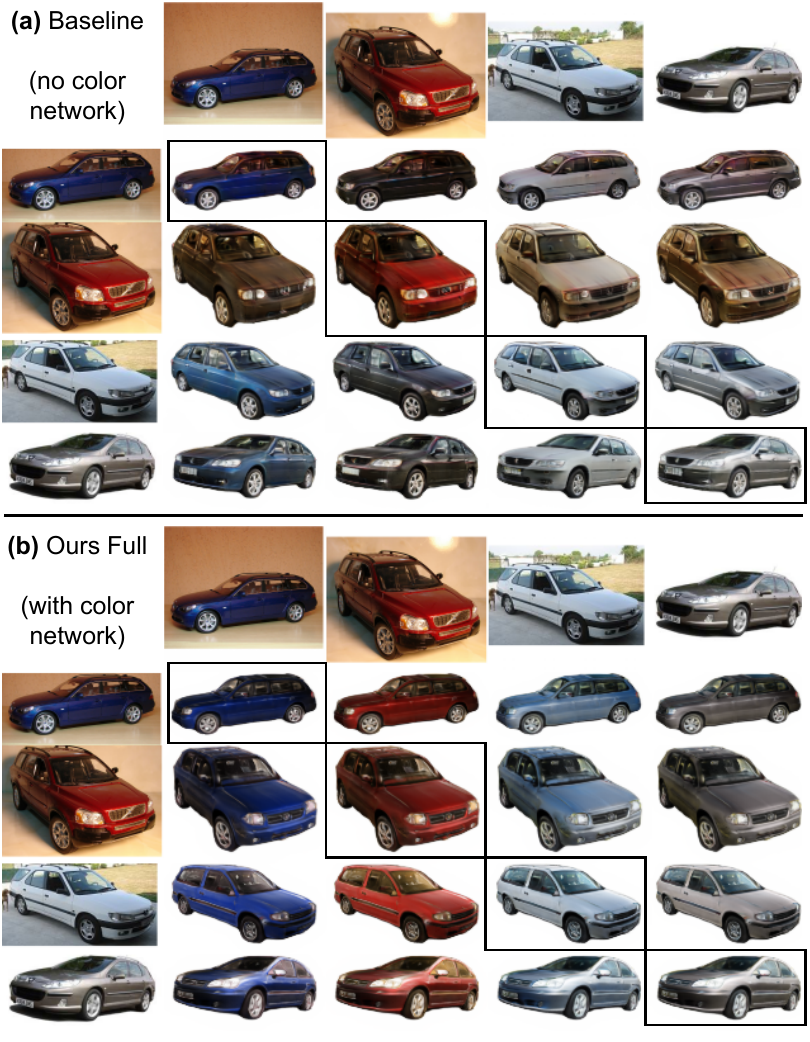}
   \caption{Disentanglement of the color distribution from the object identity. \textbf{(a}) In the baseline network without color mapping (EG3D), we attempt to achieve disentanglement via \emph{style mixing}, \ie we split the latent code $w$ into two sections (before and after the 8th layer) and mix it between the two object instances. Although this leads to some variation in color, we find that disentanglement is not properly achieved.  \textbf{(b)} With our color mapping technique, color and object identity are fully disentangled. When two different latent codes are combined, it is possible to ``borrow'' the color distribution from another image in a realistic way.}
\label{fig:appendix-color-mixing}
\end{figure}

\paragraph{Color distribution disentanglement.} To visually motivate our color mapping approach, we show examples of color disentanglement in \autoref{fig:appendix-color-mixing}. When our color mapping network is used, the object identity is fully disentangled from its color distribution. By contrast, attempting the same on a vanilla EG3D architecture is unsuccessful, even when adopting techniques such as \emph{style mixing}.

\paragraph{Pose estimation.} Pose prediction is a useful feature for AR applications and real-world datasets, where ground-truth poses are imprecise or not available. As such, it is not meant to improve performance, as it actually makes the learning task harder. Nonetheless, it is interesting to evaluate its effect on quantitative metrics. In \autoref{tab:appendix-pose-ablation}, we conduct an ablation experiment on a dataset of real images (P3D Cars) where we turn off pose prediction and use ground-truth poses from the dataset (which are imprecise). As expected, we find that qualitative metrics (FID) are mostly unaffected, whereas the IoU is degraded when switching to poses from the dataset, regardless of whether inversion is used. This confirms that, on real datasets such as P3D, individual poses are inaccurate, but our generative framework is robust to them as it relies on the overall pose distribution.
\begin{table}[ht]
\centering
\resizebox{\linewidth}{!}{
\begin{tabular}{ll|ll}
        & & \multicolumn{2}{c}{Pascal3D+ Cars} \\
 & & IoU $\uparrow$         & FID $\downarrow$ \\ \hline\hline
Ours (predicted poses) & ($N$=0) & \textbf{0.883}        & 75.90 (15.08) \\
Ours (dataset poses) & ($N$=0) & 0.803 & 73.20 (16.39) \\ \hline\hline
Ours (predicted poses) & ($N$=30) & \textbf{0.920}        & 73.53 (14.36) \\
Ours (dataset poses) & ($N$=30) & 0.802 &  72.22 (15.10)
\end{tabular}
}
\caption{Ablation experiment on pose prediction (P3D Cars dataset). The first two rows are purely feed-forward-based, while the remaining are inversion-based. In parentheses, we also report the FID on our larger test set from ImageNet.} 
\label{tab:appendix-pose-ablation}
\end{table}

\subsubsection{Qualitative results}
\label{sec:appendix-sub-qualitative}

\begin{figure}[t]
\centering
  \includegraphics[width=\linewidth, trim={0pt 3pt 0pt 3pt}, clip]{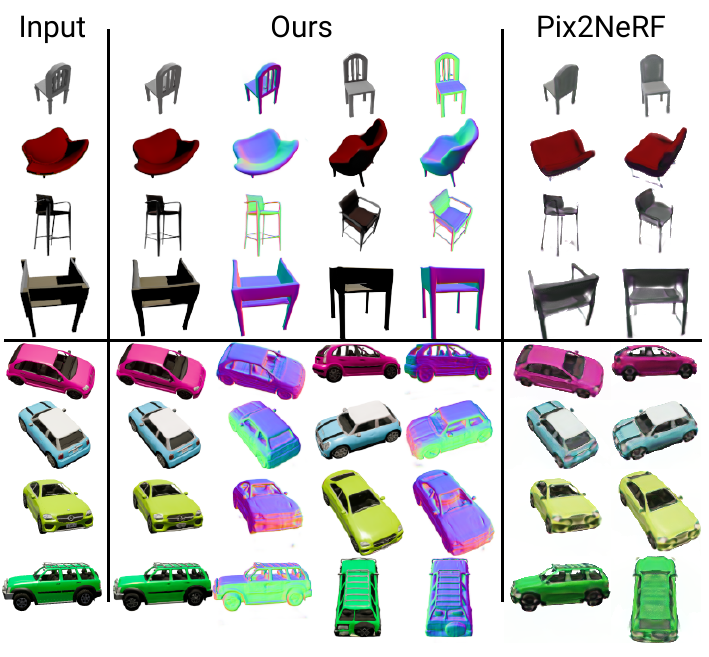}
   \caption{Additional qualitative results on synthetic datasets (test set of ShapeNet Chairs \& CARLA) and side-by-side comparison to Pix2NeRF \cite{cai2022pix2nerf} on input and random views at 128$\times$128.}
\label{fig:appendix-results-synthetic}
\end{figure}

\begin{figure*}[p]
\vspace{9mm}
\centering
  \includegraphics[width=\linewidth, trim={0pt 3pt 0pt 3pt}, clip]{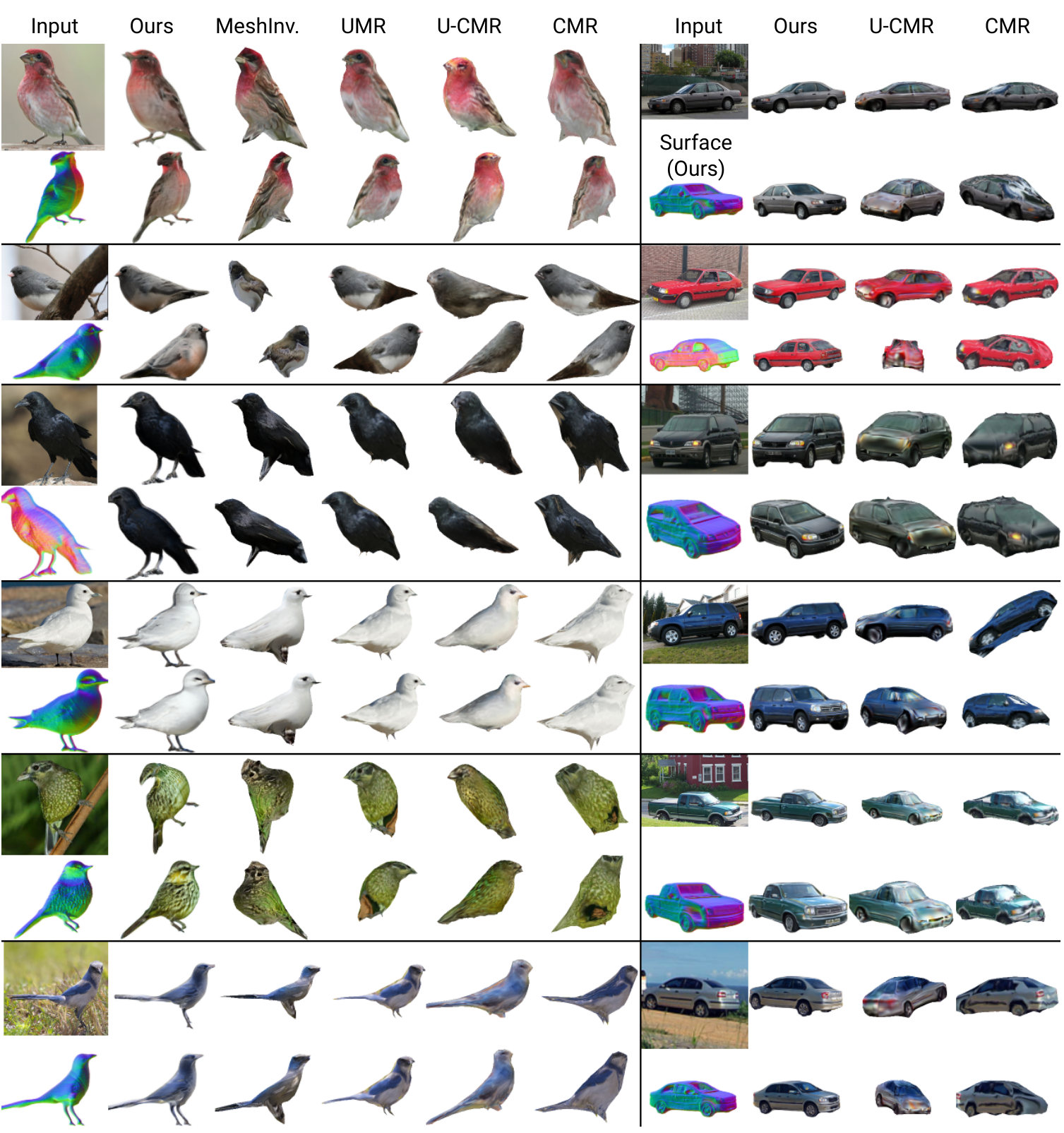}
   \caption{Additional qualitative results and side-by-side comparison on the test set of CUB (\textbf{left}) and Pascal3D+ Cars (\textbf{right}), at 128$\times$128. The first row of each sample is rendered from the input viewpoint, whereas the second row illustrates a random view.}
\label{fig:appendix-qualitative-results-p3d-cub}
\vspace{9mm}
\end{figure*}

\paragraph{Additional qualitative results.} Following the format in the main text, we report extra qualitative results for all datasets in \autoref{fig:appendix-imagenet} (novel results on ImageNet and ShapeNet Cars), \autoref{fig:appendix-results-synthetic} (ShapeNet Chairs \& CARLA, and comparison to Pix2NeRF \cite{cai2022pix2nerf}), and \autoref{fig:appendix-qualitative-results-p3d-cub} (CUB and P3D Cars, including comparison to prior work).

\begin{figure}[t]
\centering
  \includegraphics[width=\linewidth]{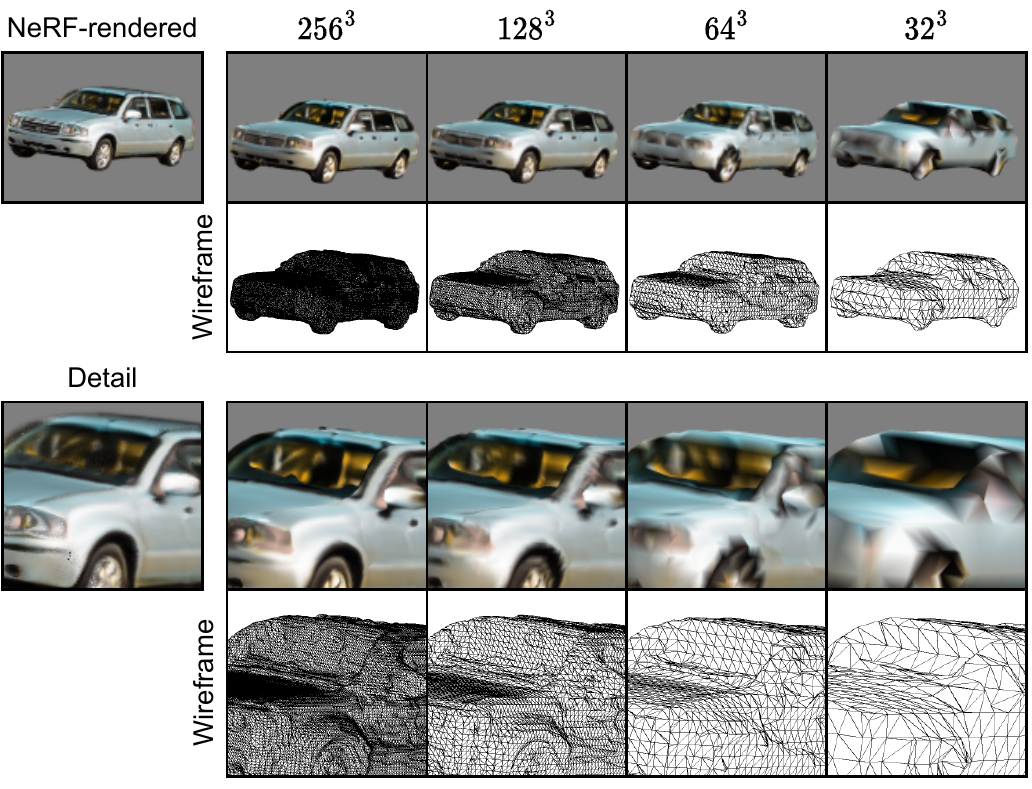}
   \caption{Extraction of a colored triangle mesh from an SDF. On the left, we show a car rendered using a neural renderer as well as zoomed viewpoint. On the right, we show the corresponding triangle mesh (including wireframe visualization) at different quantization steps.}
\label{fig:appendix-trimesh}
\end{figure}

\paragraph{Conversion to triangle mesh.} Our adoption of an SDF representation allows us to easily extract a triangle mesh from a generated object (\autoref{fig:appendix-trimesh}). We first quantize the SDF to a fixed-size grid, and then extract its 0-level set (\ie zero-crossings) via marching cubes \cite{lorensen1987marching}, obtaining a set of vertices and triangles. Finally, we sample colors from the radiance field by querying the network at the locations specified by the vertex positions.


\begin{figure}[b]
\centering
  \includegraphics[width=\linewidth]{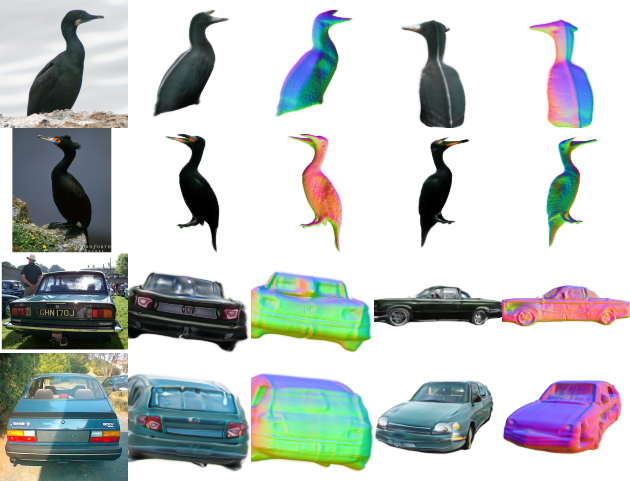}
   \caption{Most common modes of failure on CUB Birds and P3D Cars. On CUB (first two rows), we sometimes observe a ``split beak'' phenomenon. On P3D (last two rows), in the rare instances the pose is estimated imprecisely, the optimizer can get stuck in a local minima and lead to a distorted reconstruction.}
\label{fig:appendix-failure-cases}
\end{figure}

\paragraph{Demo video.} As part of the supplementary material, we include a video that shows additional examples of reconstructions. We break down each result into \emph{(i)} prediction of the canonical map, \emph{(ii)} initial pose estimation and bootstrapping of the latent code, \emph{(iii)} refinement via hybrid inversion, and \emph{(iv)} 360$^\circ$ animation of the reconstructed object and its surface. All samples are random (no selection is done) and failure cases are shown as well.

\subsubsection{Limitations and failure cases}
\label{sec:appendix-sub-failure-cases}

In this section, we highlight the most common failure modes of our method and attempt to categorize them in common patterns.

\paragraph{Shape artifacts.} In some cases, we observe that reconstructed shapes present some artifacts, even though the appearance of the object looks correct when rendered. For instance, in animals, features such as the beaks of the birds are sometimes bifurcated (\autoref{fig:appendix-failure-cases}, top). We attribute this issue to a lack of density in some areas of the pose distribution of the dataset, \eg most birds are observed from the side, but rarely from the front. On some small datasets such as zebras and elephants -- which comprise only 1.7k and 1.4k images respectively -- we also observe concavities (see \autoref{fig:qualitative-results-additional} in the main text), and in the specific case of zebras, a failure to disentangle the stripes from the shape. We expect these entanglement issues to improve with larger datasets.

\begin{figure}[ht]
\centering
  \includegraphics[width=\linewidth]{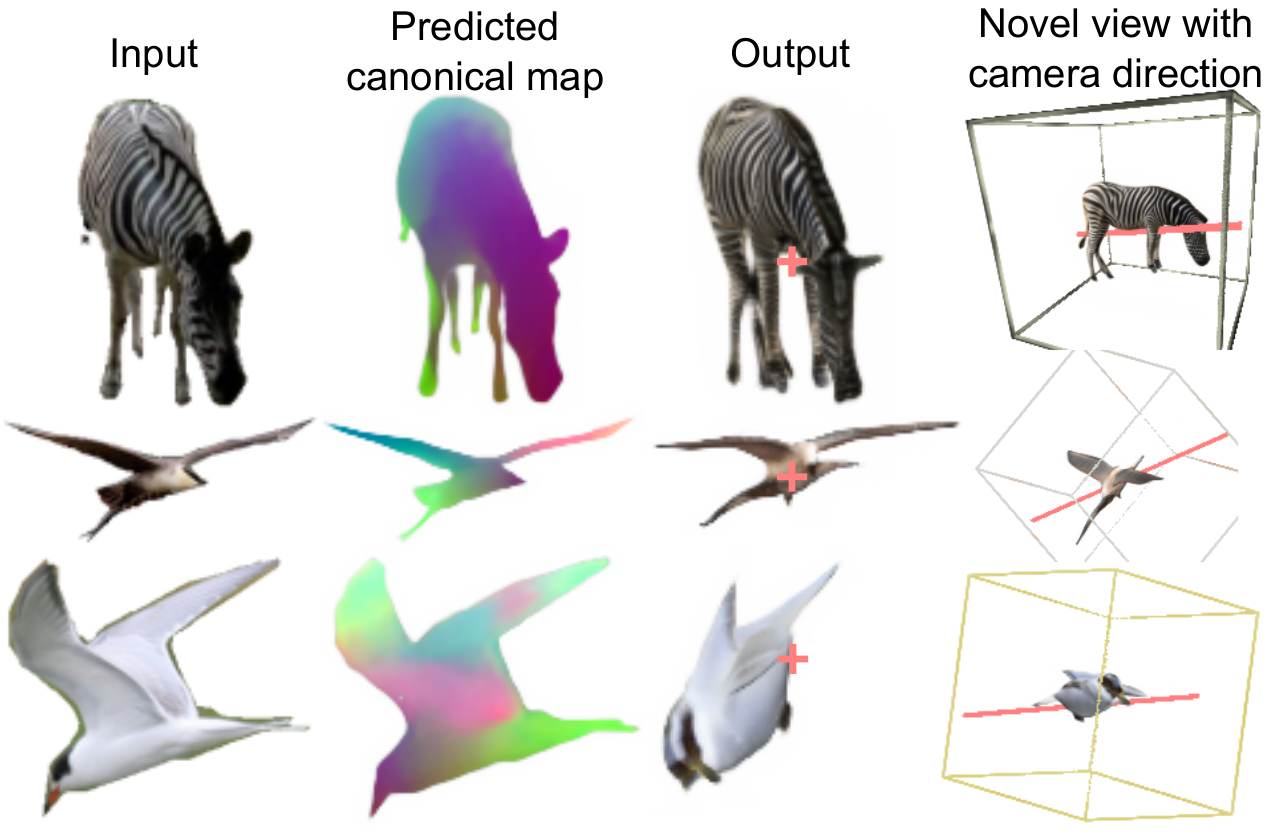}
   \caption{Our pose estimation approach can correctly handle ``extreme'' postures (top and middle), although some failure cases are still possible (bottom).}
\label{fig:appendix-pose-estimation}
\end{figure}

\paragraph{Pose estimation errors.} More rarely, failures are caused by inaccurate pose estimation. If the initial estimated pose is too far from the true one, the optimizer can get stuck in a local minima and cause the reconstructed surface to become distorted (\autoref{fig:appendix-failure-cases}, bottom). We also notice that pose estimation is more challenging on examples with ``extreme'' postures, \eg open wings in birds and head flexion in zebras, but also observe that many of these cases are handled correctly (\autoref{fig:appendix-pose-estimation}). Most likely, the failure cases are due to these poses being underrepresented in the unconditional generator, and are not due to a limitation of the pose estimation framework itself.

\newpage
\paragraph{Incomplete inversion.} For some hard examples, the encoder might return an initial solution which is far from the optimum, which in turn needs to be optimized for longer to correctly match the input image. Since we adopt a fixed schedule (\ie the same number of optimization steps for all images), some images may only be partially inverted. As part of future work, this issue could be mitigated by using an adaptive optimization schedule that varies for each sample.

\subsection{Negative results}
\label{sec:appendix-negative-results}
Throughout the development of our method, we experimented with various techniques drawing inspiration from the literature on GANs and representation learning. To guide further research in this area, we provide a list of ideas we explored but did not work out as expected.

\begin{itemize}[leftmargin=*, itemsep=-2pt]
    \item We initially experimented with various NeRF representations, including MLP-based, voxel-based, and triplanar-based. We eventually settled with the triplanar representation of \cite{chan2022eg3d} since it was as expressive as the other ones but more efficient.
    \item Our initial attempts at solving the reconstruction task used an encoder-based approach with multiple discriminators. While this was appropriate for synthetic data, we quickly found out that it was not robust on real datasets with imprecise poses, which is also one of the main motivating factors for our approach. Based on the intuition that the issue might have been caused by the limited expressivity of the encoder, we tried to replace the encoder with an embedding layer that learned a different latent code for each instance (similar to \cite{jang2021codenerf, rebain2022lolnerf}, but using a GAN framework with multiple discriminators), essentially decoupling the impact of the encoder architecture from that of the learning framework. In this setting, we found that the issue was not resolved, which prompted us to explore other ideas.
    \item Before settling with our hybrid inversion framework, we also explored techniques from the representation learning literature, such as bidirectional GANs (BiGAN) \cite{donahue2017bigan}. In this setting, the encoder is not connected to the generator and the learning signal comes from a joint discriminator. Our expectation was that this setting would make the approach less reliant on precise poses, but we found that the reconstructions did not mirror the input images closely enough.
\end{itemize}

\end{document}